\documentclass{SCIS2024}

\usepackage{graphicx}%
\usepackage{multirow}%
\usepackage{amsmath,amssymb,amsfonts}%
\usepackage{amsthm}%
\usepackage{mathrsfs}%
\usepackage[title]{appendix}%
\usepackage{xcolor}%
\usepackage{textcomp}%
\usepackage{manyfoot}%
\usepackage{booktabs}%
\usepackage{listings}%
\usepackage{makecell}
\usepackage[square, comma, sort&compress, numbers]{natbib}

\usepackage{multirow}
\usepackage{multicol}
\usepackage{enumitem}
\usepackage{hhline}
\usepackage{makecell}
\usepackage{cite}

\definecolor{lightyellow}{rgb}{1, 1, 0.93}
\begin{document}
\ArticleType{RESEARCH PAPER}
\Year{2024}
\Month{}
\Vol{}
\No{}
\DOI{}
\ArtNo{}
\ReceiveDate{}
\ReviseDate{}
\AcceptDate{}
\OnlineDate{}

\title{Large Language Models Meet Text-Centric Multimodal Sentiment Analysis: A Survey}


\author[1]{Hao YANG}{{hyang@ir.hit.edu.cn}}
\author[1]{Yanyan ZHAO}{{yyzhao@ir.hit.edu.cn}}
\author[1]{Yang WU}{}
\author[1]{Shilong WANG}{}
\author[1]{Tian ZHENG}{}
\author[1]{\\ Hongbo ZHANG}{}
\author[2]{Zongyang Ma}{}
\author[1]{Wanxiang CHE}{}
\author[1]{Bing QIN}{}

\AuthorMark{Author Hao YANG}

\AuthorCitation{Author Hao YANG, Author Yanyan ZHAO, et al}


\address[1]{Harbin Institute of Technology, Harbin {\rm 150001}, China}
\address[2]{Chinese Academy of Sciences, Beijing {\rm 100190}, China}

\abstract{

Compared to traditional sentiment analysis, which only considers text, multimodal sentiment analysis needs to consider emotional signals from multimodal sources simultaneously and is therefore more consistent with the way how humans process sentiment in real-world scenarios. It involves processing emotional information from various sources such as natural language, images, videos, audio, physiological signals, etc. However, although other modalities also contain diverse emotional cues, natural language usually contains richer contextual information and therefore always occupies a crucial position in multimodal sentiment analysis. The emergence of ChatGPT has opened up immense potential for applying large language models (LLMs) to text-centric multimodal tasks. However, it is still unclear how existing LLMs can adapt better to text-centric multimodal sentiment analysis tasks. This survey aims to (1) present a comprehensive review of recent research in text-centric multimodal sentiment analysis tasks, (2) examine the potential of LLMs for text-centric multimodal sentiment analysis, outlining their approaches, advantages, and limitations, (3) summarize the application scenarios of LLM-based multimodal sentiment analysis technology, and (4) explore the challenges and potential research directions for multimodal sentiment analysis in the future.
}

\maketitle

\section{Introduction}

Text-based sentiment analysis is a crucial research task in the field of natural language processing, aiming at automatically uncovering the underlying attitude that we hold towards textual content. However, humans often process emotions in a multi-modal environment, which differs from text-based scenarios in the following ways: 

\textbf{1) Humans have the ability to acquire and integrate multimodal fine-grained signals.} Humans often find themselves in multimodal scenarios, manifested as seamlessly understanding others' intentions and emotions through the combined effects of language, images, sound, and physiological signals. When processing emotions, humans have the ability to sensitively capture and integrate fine-grained sentiment signals from multiple modalities, and correlate them for emotional reasoning. 

\textbf{2) Multimodal expression ability.} The ways in which humans express emotions include language, facial expressions, body movements, speech, etc. For example, in daily conversations, our natural language expressions may be vague (such as someone saying ``okay''), but when combined with other modal information, like visual modalities (e.g. a happy facial expression) or audio modalities (e.g. a prolonged intonation), the emotions expressed are different.

It is evident that the study of sentiment analysis within a multimodal context brings us closer to authentic human emotion processing. Research into multimodal sentiment analysis technologies \cite{5,109} with human-like emotion processing capabilities will provide technical support for real-world applications such as high-quality intelligent companions, customer service, e-commerce, and depression detection. In recent years, large language models (LLMs) \cite{2,3,4} have demonstrated astonishing human-machine conversational capabilities and showcased impressive performance across a wide range of natural language processing tasks, indicating their rich knowledge and powerful reasoning abilities. At the same time, large multimodal models (LMM) that increase the ability to understand modalities such as images also provide new ideas for multimodal-related tasks. They can directly perform tasks with zero-shot or few-shot context learning, requiring no supervised training \cite{7,10,11,12}. While there have been some attempts to apply LLMs in text-based sentiment analysis \cite{6,7,8,9,108}, there is a lack of systematic and comprehensive analysis regarding the application of LLMs and LMMs in multimodal sentiment analysis. Therefore, it remains unclear to what extent existing LLMs and LMMs can be used for multimodal sentiment analysis.

Given the crucial role of natural language in multi-modal sentiment analysis and its essential input for current LLMs and LMMs, we concentrate on text-centric multimodal sentiment analysis tasks that can leverage LLMs to enhance performance, such as image-text sentiment classification, image-text emotion classification, audio-image-text (video) sentiment classification, etc. In this work, we aim to provide a comprehensive review of the current state of text-centric multimodal sentiment analysis methods based on LLMs and LMMs. Specifically, we focus on the following questions: \textit{1) How do LLMs and LMMs perform in a variety of multimodal sentiment analysis tasks? 2) What are the differences among approaches to utilize LLMs and LMMs in various multimodal sentiment analysis tasks, and what are their respective strengths and limitations? 3) What are the future application scenarios of multimodal sentiment analysis?}

\begin{figure*}
    \centering
    \includegraphics[width=16cm]{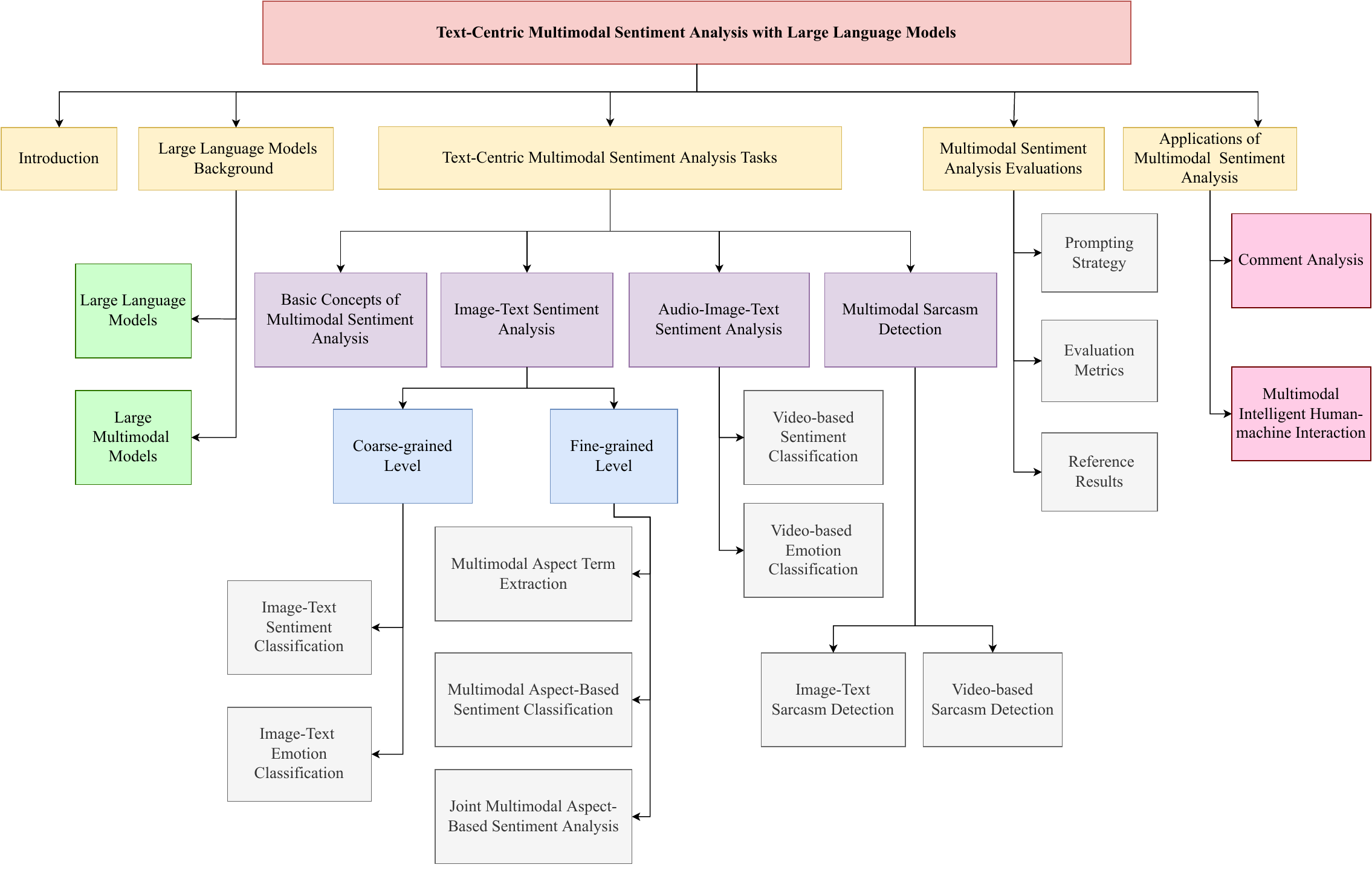}
    \caption{Organization of the review article.}
    \label{fig:all}
\end{figure*}

To this end, we first introduce the tasks and the most recent advancements in text-centric multimodal sentiment analysis. We also outline the primary challenges faced by current technologies and propose potential solutions. We examine a total of 14 multimodal sentiment analysis tasks, which have traditionally been studied independently. We analyze the distinct characteristics and commonalities of each task. The structure of the review study is depicted in Figure~\ref{fig:all}. Since LMMs are also based on LLMs, for convenience of presentation, the methods based on LLMs below include methods based on LMMs.

The rest of the sections of this paper are organized as follows. Section \ref{section2} introduces the background of LLMs and LMMs. In Section~\ref{section3}, we conduct an extensive survey on a wide range of text-centric multimodal sentiment analysis tasks, detailing the task definitions, related datasets and the latest methods. We also summarize the advantages and advancements of LLM compared to previous techniques in multimodal sentiment analysis tasks, as well as the challenges still faced. In Section~\ref{section4}, We introduced the prompt settings, evaluation metrics, and reference results related to LLMs-based text-centric multimodal sentiment analysis methods.  In Section~\ref{section5}, we look forward to the future application scenarios of multimodal sentiment analysis, followed by concluding remarks in Section~\ref{section6}.

\section{Background on Large Language Models}\label{section2}

\subsection{Large Language Models}
Generally, LLMs refer to transformer models with hundreds of billions (or more) of parameters, which are trained on large amounts of text data at a high cost, such as GPT-3 \cite{2}, PaLM \cite{22}, Galactica \cite{23}, and LLaMA2 \cite{24}. LLMs typically possess extensive knowledge and demonstrate strong abilities in understanding, generating natural language, and solving complex tasks in practical. LLMs exhibit some abilities that are not present in small models, which is the most prominent feature that distinguishes LLM from previous pre-trained language models(PLMs), for example, in-context learning (ICL) capacity. Assuming that the language model has been provided with natural language instructions and several task demonstrations, it can generate the expected output of the test instance by completing the word sequence of the input text without additional training or gradient updates; \textit{Instruction following.} By fine-tuning the mixture of multi-task datasets formatted through natural language descriptions (known as instruction adaptation), LLM performs well on unseen tasks also described in instruction form. Through fine-tuning instructions, LLM is able to follow task instructions for new tasks without using explicit examples, thus improving generalization ability. \textit{Step-by-step reasoning.} For small language models(SLMs), it is often difficult to solve complex tasks involving multiple reasoning steps, such as mathematical word problems. Instead, using the chain-of-thought (CoT) cueing strategy \cite{25,26,27}, LLMs can solve such tasks by leveraging a cueing mechanism that involves intermediate reasoning steps to derive the final answer. 

There have been some preliminary attempts to evaluate LLMs for text sentiment analysis tasks. In \cite{7}, the authors observed that the zero-shot performance of LLMs can be compared with fine-tuning BERT models \cite{105}. In addition, in \cite{8}, the authors conducted preliminary research on some sentiment analysis tasks using ChatGPT, specifically studying its ability to handle polarity changes, open-domain scenarios, and emotional reasoning problems. In \cite{9}, the authors comprehensively tested the effectiveness of LLMs in text sentiment analysis datasets. In \cite{28}, the authors tested the effectiveness of commercial LLMs on a multimodal video-based sentiment analysis dataset. Despite these existing efforts, their scope is often limited to partial tasks and involves different datasets and experimental designs. Our goal is to comprehensively summarize the performance of LLMs in the field of multimodal sentiment analysis.

\subsection{Large Multimodal Models}
Large multimodal models (LMMs) are created to handle and integrate various data types, such as text, images, audio, and video. LMMs extend the capabilities of LLMs by incorporating additional modalities, allowing for a more comprehensive understanding and generation of diverse content. The development of LMMs is driven by the need to more accurately reflect the multimodal nature of human communication and perception. While traditional LLMs like GPT-4 are primarily text-based, LMMs are capable of processing and generating outputs across various data types. For instance, they can interpret visual inputs, generate textual descriptions from images, and even handle audio data, thus bridging the gap between different forms of information. One of the critical advancements in LMMs is the ability to create a unified multimodal embedding space. This involves using separate encoders for each modality to generate data-specific representations, which are then aligned into a cohesive multimodal space. This unified approach allows the models to integrate and correlate information from different sources seamlessly. 

Notable examples include Gemini \cite{111}, GPT-4V, and ImageBind \cite{110}. These models showcase the ability to process text, images, audio, and video, enhancing functionalities such as translation, image recognition, and more. 
In addition to these well-known models, other emerging models are also making significant strides: 
BLIP-2 \cite{112} introduces a novel approach to integrate a frozen pre-trained visual encoder with a frozen large language model using a Q-former module. This module employs learnable input queries that interact with image features and the LLM, allowing for effective cross-modal learning. This setup helps maintain the versatility of the LLM while incorporating visual information effectively. LLava \cite{113} is a represent large multimodal model integrating a pre-trained CLIP \cite{116} visual encoder (ViT-L/14), the Vicuna \cite{115} language model, and a simple linear projection layer. Its training involves two stages: feature alignment pre-training, where only the projection layer is trained using 595K image-text pairs from Conceptual Captions dataset \cite{118}, and end-to-end fine-tuning, where the projection layer and LLM are fine-tuned using 158K instruction-following data and the ScienceQA dataset \cite{117}. This setup ensures effective integration of visual and textual information, enabling LLava to excel in image captioning, visual question answering, and visual reasoning tasks. Qwen-VL \cite{114} is a strong performer in the multimodal domain. Qwen-VL excels in tasks such as zero-shot image captioning and visual question answering, supporting both English and Chinese text recognition. Qwen-VL-Chat enhances interaction capabilities with multi-image inputs and multi-round question answering, showcasing significant improvements in understanding and generating multimodal content.

\subsection{Usage of Large Language Models}
In~\cite{208}, the authors summarize two paradigms for utilizing LLMs: \textbf{Parameter-frozen} application directly applies prompting approach on LLMs without the need for parameter tuning. This category includes zero-shot and few-shot learning, depending on whether the few-shot demonstrations is required. \textbf{Parameter-tuning} application refers to the need for tuning parameters of LLMs. This category includes both full-parameter and parameter-efficient tuning, depending on whether fine-tuning is required for all model parameters.

In zero-shot learning, LLMs leverage the instruction following capabilities to solve downstream tasks based on a given instruction prompt, which is defined as:
\begin{gather}
    P = Prompt(I), 
\end{gather}
where $I$ and $P$ denote the input and output of prompting, respectively.

Few-shot learning uses in-context learning capabilities to solve the downstream tasks imitating few-shot demonstrations. Formally, given some demonstrations $E$, the process of few-shot learning is defined as:
\begin{gather}
    P = Prompt(E,I).
\end{gather}

In the full-parameter tuning approach, all parameters of the model $M$ are fine-tuned on the training dataset $D$:
\begin{gather}
    \widehat{M} = Fine\text{-}tune(M|D), 
\end{gather}

where $\widehat{M}$ is the fine-tuned model with the updated parameters.

Parameter-efficient tuning (PET) involves adjusting a set of existing parameters or incorporating additional tunable parameters (like Bottleneck Adapter~\cite{209}, Low-Rank Adaptation (LoRA)~\cite{210}, Prefix-tuning~\cite{211}, and QLoRA~\cite{212}) to efficiently adapt models for specific downstream tasks. Formally, parameter-efficient tuning first tunes a set of parameters $W$, denoting as:
\begin{gather}
    \widehat{W} = Fine\text{-}tune(W|D, M),
\end{gather}
where $\widehat{W}$ stands for the trained parameters.

\section{Text-Centric Multimodal Sentiment Analysis Tasks }\label{section3}
Text-centered multimodal sentiment analysis mainly includes image-text sentiment analysis and audio-image-text (video) sentiment analysis. Among them, according to different emotional annotations, the two most common tasks are sentiment classification tasks (such as the most common three label classification tasks of positive, neutral, and negative) and emotion classification tasks (including emotional labels such as happy, sad, angry, etc). Similar to text-based sentiment classification, text-centered multimodal sentiment analysis can also be categorized into coarse-grained multimodal sentiment analysis (e.g., sentence-level) and fine-grained multimodal sentiment analysis (e.g., aspect-level) based on the granularity of the opinion targets. Existing fine-grained multimodal sentiment analysis usually focuses on image-text pair data, and includes multimodal aspect term extraction (MATE), multimodal aspect-based sentiment classification (MASC), and joint multimodal aspect-sentiment analysis (JMASA). Additionally, multimodal sarcasm detection has also become a widely discussed task in recent years. Due to the need to analyze conflicts between different modalities of sentiment, it highlights the importance of non-text modalities in sentiment judgment in real-world scenarios. We will introduce these tasks in the following subsections, and summarize them in Table~\ref{tab:multimodal works}.

\setcounter{table}{0}
\begin{table*} [!htb]
\centering\small
\caption{Categorization and representative methods for text-centric multimodal sentiment analysis.}
\scalebox{0.65}{
\colorbox{lightyellow}{

\begin{tabular}{c|cc|c|c}
\hline
Category                    & \multicolumn{2}{c|}{Task}                                                                               & Datasets                                                                                  & Methods                                \\ \hline
\multirow{6}{*}{Image-Text} & \multicolumn{1}{c|}{\multirow{3}{*}{Coarse-grained}} & Image-Text Sentiment Classification              & MVSA~\cite{121}, MEMOTION 2~\cite{123}, MSED~\cite{140}                                                                          & \cite{31,130,131,132,133,134,135,136,137,168,169}                      \\ \cline{3-5} 
                            & \multicolumn{1}{c|}{}                                & Image-Text Emotion Classification                & TumEoM~\cite{127}, MEMOTION 2~\cite{123}, MSED~\cite{140}                                                                                     &  \cite{127,137,139,168,169}                                        \\ \cline{3-5} 
                            & \multicolumn{1}{c|}{}                                & Image-Text Sarcasm Detection                     &  MMSD~\cite{120}, MMSD2.0~\cite{205}                                                                         & \cite{55,194,195,196,197,198,199,200,202,203,205,206,207}    \\ \cline{2-5} 
                            & \multicolumn{1}{c|}{\multirow{3}{*}{Fine-grained}}   & Multimodal Aspect Term Extraction                & Twitter-15~\cite{79}, Twitter-17~\cite{79}                                                                    & \cite{96,145,146,147,148,149,150}       \\ \cline{3-5} 
                            & \multicolumn{1}{c|}{}                                & Multimodal Aspect Sentiment Classification & Multi-ZOL~\cite{77}, Twitter-15~\cite{79}, Twitter-17~\cite{79}                                                         & \cite{32,77,78,79,98,160,168,169}                         \\ \cline{3-5} 
                            & \multicolumn{1}{c|}{}                                & Joint Multimodal Aspect-Sentiment Analysis       & Twitter-15~\cite{79}, Twitter-17~\cite{79}                                                                    & \cite{80,161,162,163,164,165,166,167,170,171}                               \\ \hline
\multirow{3}{*}{Video}      & \multicolumn{2}{c|}{Video-based Sentiment Classification}                                               & \begin{tabular}[c]{@{}c@{}}ICT-MMMO~\cite{193},CMU-MOSI~\cite{99},\\ CMU-MOSEI~\cite{100}, CMU-MOSEAS~\cite{125}, \\CH-SIMS~\cite{101}, CH-SIMS 2~\cite{173}, MELD~\cite{122}\end{tabular} & \cite{33,34,35,36,37,38,39,40,41,42,43,181,182,183,184,185,186,187,188,189,190,191,192,204} \\ \cline{2-5} 
                            & \multicolumn{2}{c|}{Video-based Emotion Classification}                                                 & \begin{tabular}[c]{@{}c@{}}MELD~\cite{122},IEMOCAP~\cite{126},CMU-MOSEI~\cite{100},\\ M3ED~\cite{174},MER2023~\cite{175},\\EMER~\cite{177},ER2024~\cite{176}\end{tabular}                                                              & \begin{tabular}[c]{@{}c@{}}\cite{122,126,174,175,176,177,184}\\ \cite{186,187,188,190,192}\end{tabular} 
                                                \\ \cline{2-5} 
                            & \multicolumn{2}{c|}{Video-based Sarcasm Detection}                                                      & MUStARD~\cite{124}                                                                                   & \cite{53,54,57}    \\ \hline
\end{tabular}

}
}
\label{tab:multimodal works} 
\end{table*}

\subsection{Basic Concepts of Multimodal Sentiment Analysis}
Multimodal sentiment analysis (MSA) differs from traditional text-based sentiment analysis in that it combines multiple modalities, such as images and speech, to enhance the accuracy of sentiment classification. The most common multimodal sentiment analysis scenarios include ``image-text'', ``audio-image'' and ``audio-image-text'' (video). For example, the sentence "That's great!" expresses a positive emotion when analyzed as text alone, but when combined with an eye-rolling expression and a sharp tone of voice, the overall sentiment is sarcastically negative. Additionally, multimodal scenarios can also extend to more modalities  that can reflect human emotions, such as ``physiological signals'' (skin conductance, electromyography, blood pressure, electroencephalography, respiration, pulse, electrocardiogram, etc.). In the following chapters of this paper, we will primarily focus on key tasks and techniques for text-centric multimodal sentiment analysis in ``image-text'' and ``audio-image-text'' (video) scenarios that can leverage large language models (LLMs). Since the ``physiological signals'' modality is interdisciplinary, encompassing fields like neuroscience and psychology, and has wide-ranging application potential, we will also provide a brief overview of it.

Although multimodal data contains richer information, effectively integrating multimodal information is a key challenge in current multimodal sentiment analysis tasks. Unlike sentiment expression in text-only modalities, sentiment expression in a multimodal context has its own particularities, including: 1) Complexity of sentiment semantic representation. In multimodal scenarios, sentiment semantics are derived from the representations of each participating modality. However, each single modality can have various representation methods, making the selection of which representation to use and how to fuse the representations from multiple modalities complex. 2) Complementarity of sentiment elements. Due to the participation of other modalities, the textual modality often has shorter and less informative expressions. Fine-grained sentiment elements from other modalities can provide effective supplements. 3) Inconsistency in sentiment expression: There can be conflicts in sentiment expressions among different modalities in the same scenario, with irony being the most common example. 

Therefore, the core of multimodal sentiment analysis includes independent representation of single-modal sentiment semantics and fusion of multimodal sentiment semantic representations.

Independent representation of multimodal semantics refers to encoding each type of modality data separately. The encoding for each modality may take different forms and may not exist in the same semantic space. With the development of deep learning, deep learning techniques have shown outstanding performance in fields such as natural language processing, computer vision, and speech recognition. One of the greatest advantages is that many deep learning models (such as CNN \cite{85}: Convolutional Neural Network) and concepts can be used across these three research areas, significantly lowering the research threshold for researchers and breaking down the barriers to joint representation of multimodal semantics. Each modality can be represented as vector information through deep learning models, and simple vector concatenation and addition can achieve the most basic multimodal semantic fusion, which serves as the basis for completing other multimodal downstream tasks. Additionally, researchers have found that each modality's representation is an independent modality space representation, residing in different vector spaces. Although rigid concatenation and addition have shown some effects, their theoretical significance is hard to justify. Therefore, scholars have begun to think about how to unify multiple modality representations into the same semantic space. For example, CLIP \cite{116} uses techniques like contrastive learning and pre-training to obtain unified representations of images and text. This unified representation of multimodal semantics is also referred to as multimodal semantic fusion.

The fusion of multimodal sentiment semantic representation typically includes feature layer fusion, algorithm layer fusion, and decision layer fusion \cite{119}.  Feature layer Fusion (Early Fusion). This refers to the straightforward method of feature concatenation directly after extracting features from each modality. Algorithm Layer Fusion (Model-level Fusion). This refers to thoroughly integrating each modality within different algorithmic frameworks. For example, two modalities can undergo nonlinear transformations through their respective deep learning models to achieve more abstract representations, sharing the same loss function to achieve comprehensive modality fusion. Decision Layer Fusion (Late Fusion). This refers to combining each modality's representations with specific classification tasks to obtain independent representations for each modality and then using these to make the final classification decision. These approaches aim to address how to eliminate conflicts between modalities and how to achieve information complementarity among them.

\setcounter{table}{1}
\tabcolsep 9pt
\renewcommand\arraystretch{1.3}
\begin{table*}[!htb]
\centering
\caption{Datasets of text-centric multimodal sentiment analysis task. We use 'Emotions' to indicate that the dataset includes emotional labels, for example, happy, surprise, sad, angry, etc. And numeric intervals to represent the sentiment scoring annotations of the dataset.}
{\footnotesize
\scalebox{0.87}{
\begin{tabular}{ccccccc}
\\ \hline
Dataset & Language & Source             & year & Size   & Modalities & Labels                                                                 \\ \hline
ICT-MMMO~\cite{172} & English & YouTube & 2011 & 340 & A+V+T & [-2,2] \\
IEMOCAP~\cite{126} & English & Shows & 2008 & 10,039 & A+V+T & Emotions \\
CMU-MOSI~\cite{99}    & English  & YouTube            & 2016        & 2,199           & A+V+T & Neg, Neu, Pos                                                    \\
CMU-MOSEI~\cite{100}   & English  & YouTube            & 2018      & 23,453           & A+V+T & Neg, Neu, Pos and Emotions                                                    \\
MELD~\cite{122} & English & Movies, TVs & 2019 & 1,443 & A+V+T & Neg, Neu, Pos and Emotions \\
CH-SIMS~\cite{101} & Chinese  & Movies, TVs & 2020       & 2,281           & A+V+T  & [-1,1]           \\
CH-SIMS 2~\cite{173} & Chinese  & Movies, TVs & 2022       & 4,406           & A+V+T  & [-1,1]           \\
M3ED~\cite{174} & Chinese  & Movies, TVs & 2022       & 24,449           & A+V+T  & Emotions           \\
MER2023~\cite{175} & Chinese  & Movies, TVs & 2023       & 3,784           & A+V+T  & Emotions           \\
EMER~\cite{177} & Chinese  & Movies, TVs & 2023       & 100           & A+V+T  & Emotions, Reasoning           \\
MER2024~\cite{176} & Chinese  & Movies, TVs & 2024       & 6,199           & A+V+T  & Emotions           \\
CMU-MOSEAS~\cite{125} & \begin{tabular}[c]{@{}c@{}}Spanish, Portuguese, \\ German, French\end{tabular}  & YouTube            & 2021      & 40,000           & A+V+T  & [-3,3], [0,3]                                                   \\
UR-FUNNY~\cite{178} & English  & Speech Video & 2023       & 16,514           & A+V+T  & Funny           \\
TumEoM~\cite{127} & English & Tumblr & 2020 & 195,264 & V+T & Emotions \\
MVSA~\cite{121} & English & Twitter & 2021 & 19,598 & V+T & Neg, Neu, Pos\\
Multi-ZOL~\cite{77} & Chinese & ZOL.com & 2019 & 5,288 & V+T & [1,10]\\
MEMOTION 2~\cite{123} & English & Reddit, Facebook & 2022 & 10,000 & V+T & Neg, Neu, Pos \\
MSED~\cite{140} & English & \begin{tabular}[c]{@{}c@{}}Getty Image, Flickr \\ and Twitter\end{tabular}
 & 2022 & 9,190 & V+T & Neg, Neu, Pos and Emotions \\
Twitter-2015~\cite{79} & English & Twitter & 2019 & 5,338 & V+T & Neg, Neu, Pos \\
Twitter-2017~\cite{79} & English & Twitter & 2019 & 5,972 & V+T & Neg, Neu, Pos \\
MMSD~\cite{120} & English & Twitter & 2019 & 24,635 & V+T & Neg, Pos\\
MMSD2.0~\cite{205} & English & Twitter & 2023 & 24,635 & V+T & Neg, Pos\\
MUStARD~\cite{124} & English & Movies, TVs & 2021 & 690 & A+V+T & Neg, Pos
\\ \hline
\end{tabular}
}
}
\end{table*}

\subsection{Image-Text Sentiment Analysis}
\subsubsection{Coarse-grained Level}
Image-text coarse-grained sentiment analysis primarily encompasses two tasks: emotion classification and sentiment classification. Given an image-text pair, the emotion classification task aims to identify emotional labels such as happiness, sadness, surprise, etc, inspired by the text-based emotion classification task. Sentiment classification aims to identify the sentiment label, which usually includes three categories (positive, neutral, negative). Problem formalization as follows: 

Given a set of multimodal posts from social media, $P = \left\{ (T_{1}, V_{1}), ..., (T_{N}, V_{N}) \right\}$, where $T_{i}$ is the text modality and $V_{i}$ is the corresponding visual information, $N$ represents the number of posts. We need to learn the model $f : P \xrightarrow{} L$ to classify each post $(T_{i}, V_{i})$ into the predefined categories $L_{i}$. For polarity classification, $L_{i} \in \left\{ Positive, Neutral, Negative\right\}$; for emotion classification, $L_{i} \in \left\{ Angry, Bored, Calm, \right.$ $\left. Fear, Happy, Love, Sad \right\}$.

The earliest image-text sentiment classification models were feature-based. In~\cite{130}, the authors used SentiBank to extract 1200 adjective-noun pairs (ANPs) as visual semantic features and employed SentiStrength~\cite{129} to compute text sentiment features for handling multimodal tweet sentiment analysis. In~\cite{131}, the authors presented a cross-media bag-of-words model to represent the text and image of a Weibo tweet as a unified bag-of-words representation. Then some neural network models showed better performance. In~\cite{132,133}, the authors used convolutional neural network (CNN) models to get the representation of text and image. In~\cite{134}, the authors believed that more detailed semantic information in the image is important and constructed HSAN, a hierarchical semantic attentional network based on image caption for coarse-level multimodal sentiment analysis. MultiSentiNet~\cite{135} focused on the correlation between images and text, aggregating the representation of informative words with visual semantic features, objects, and scenes. Considering the mutual influence between image and text, Co-Mem~\cite{136} is designed to iteratively model the interactions between visual contents and textual words for multimodal sentiment analysis.

In~\cite{31}, the authors found that images play a supporting role to text in many sentiment detection cases, and proposed VistaNet which instead of using visual information as features only rely on visual information as alignment for pointing out the important sentences of a document using attention. With respect to each image representation $f^{v}_{j}$ , the goal is to learn the attention weights $\beta_{j,i}$ for text representations $f^{t}_{j}$:
\begin{gather}
    p_{j} = \tanh (W_{p}f^{v}_{j}+b_{p}), \\
    q_{i} = \tanh (W_{q}f^{t}_{i}+b_{q}), \\
    v_{j,i} = V^{T}(p_{j}\odot q_{i} + q_{i}), \\
    \beta_{j,i} = \frac{\exp(v_{j,i})}{\sum_{i}\exp(v_{j,i})}
\end{gather}
where $W_{p}, W_{q}, b_{p}, b_{q}$ are learnable parameters, firstly, projecting both image representation and text representation onto
an attention space followed by a non-linear activation function $\tanh$. Then, let the image projection $p_{j}$ interact with the
sentence projection $q_{i}$ in two ways: element-wise multiplication and summation.

CLMLF~\cite{138} applies contrastive learning and data augmentation to align and fuse the token-level features of text and image. In addition to focusing on sentiment, emotions are equally important. In~\cite{127}, the authors built an image-text emotion dataset, named TumEmo, and further proposed MVAN for multi-modal emotion analysis. In~\cite{137}, the authors observed that multimodal emotion expressions have specific global features and introduced a graph neural network, proposing an emotion-aware multichannel graph neural network method called MGNNS. MULSER~\cite{139} is also a graph-based fusion method that not only investigates the semantic relationship among objects and words respectively, but also explores the semantic relationship between regional objects and global concepts, which has also yielded effective results.

\begin{figure*}[h]
    \centering
    \includegraphics[width=16cm]{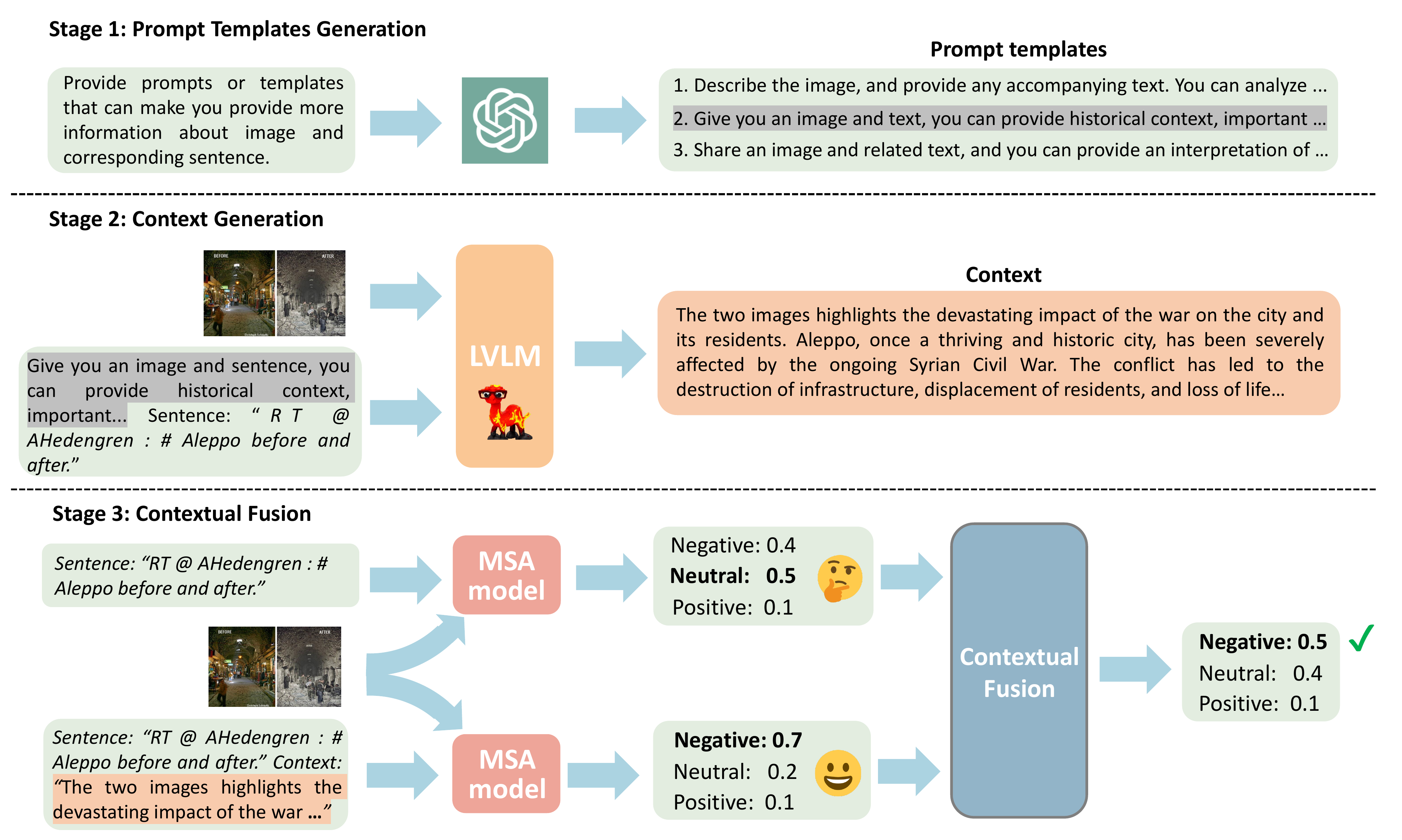}
    \caption{Overview of WisdoM framework~\cite{171} architecture for  coarse-grained image-text sentimen.}
    \label{fig:cits}
\end{figure*}

Traditional multimodal sentiment analysis methods often rely on superficial information, lacking the depth provided by contextual world knowledge. This limits their ability to accurately interpret the sentiment conveyed in multimodal content (images and text). WisdoM~\cite{141} leverage the contextual world knowledge induced from the LMMs for enhanced multimodal sentiment analysis. As shown in Figure~\ref{fig:cits}, the process involves three stages:
1) Prompt Templates Generation: Using ChatGPT to create templates that help LMMs understand the context better.
2) Context Generation: Feeding these templates into LMMs along with the sentence and image to generate rich contextual information. 3) Contextual Fusion: Combining this contextual information with the original sentiment predictions to enhance accuracy, particularly for difficult samples. A training-free module called Contextual Fusion is introduced to minimize noise in the contextual data, ensuring that only relevant information is considered during sentiment analysis. WisdoM significantly outperforms existing state-of-the-art methods in MSED dataset, demonstrating its effectiveness in integrating contextual knowledge for improved sentiment classification.
In addition, inspired by the success of textual prompt-based fine-tuning approaches in few-shot scenario, the authors~\cite{168} introduce a multi-modal prompt-based fine-tuning approach UP-MPF and the authors~\cite{169} propose a prompt-based vision-aware language modeling (PVLM) for multimodal sentiment analysis.

We summarize the commonly used datasets for coarse-grained image-text sentiment and emotion analysis, including TumEom, MVSA, MEMOTION 2, and MSED:

\textbf{TumEmo} is a multimodal weak-supervision emotion dataset containing a large amount of image-text data crawled from Tumblr. The dataset contains 195,265 image-text pairs with 7 emotion labels: Angry, Bored, Calm, Fearful, Happy, Loving, Sad.

\textbf{MVSA} dataset is collected from image-text pairs on the Twitter platform and is manually annotated with three sentiment labels: positive, neutral, and negative. The MVSA dataset consists of two parts: MVSA-Single, where each sample is annotated by a single annotator, comprising 4,869 image-text pairs, and MVSA-Multiple, where each sample is annotated by three annotators with three emotion labels, totaling 19,598 image-text pairs. The MVSA corpus is another example of coarse-grained multimodal sentiment classification dataset.

\textbf{MEMOTION 2} is a dataset focused on classifying emotions and their intensities into discrete labels. It includes 10,000 memes collected from various social media sites. These memes are typically humorous and aim to evoke a response. Overall Sentiment (positive, neutral, negative), Emotion (humour, sarcasm, offence, motivation), and Scale of Emotion are all annotated for each meme (0–4 levels).

\textbf{MSED} comprises 9,190 pairs of text and images sourced from diverse social media platforms, including but not limited to Twitter, Getty Images, and Flickr. Each piece of multi-modal sample is manually annotated with desire category, sentiment
category (i.e., positive, neutral and negative) and emotion category (happiness, sad, neutral, disgust, anger and fear).

\begin{figure*}[h]
    \centering
    \includegraphics[width=16cm]{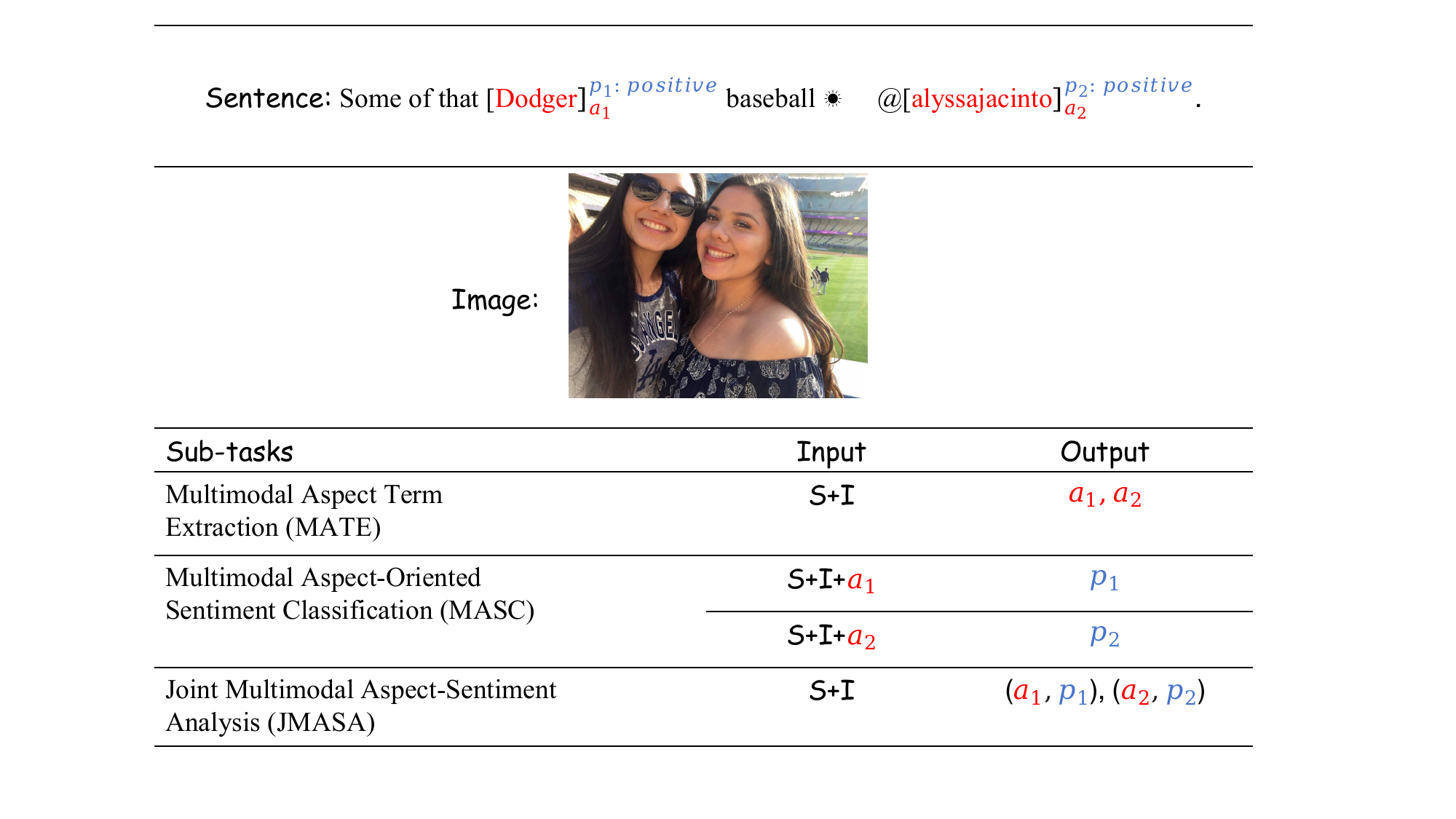}
    \caption{Image-text fine-grained sentiment analysis tasks.}
    \label{fig:absa}
\end{figure*}

\subsubsection{Fine-grained Level}
Image-text fine-grained sentiment analysis focuses on analyzing sentiment elements that are finer than sentence-level, such as \color{red}{aspect term (a)} \color{black}{and} \color{blue}{sentiment polarity (p)} \color{black}, or their combinations. It has received widespread attention in recent years and mainly includes three subtasks: Multimodal Aspect Term Extraction (MATE), Multimodal Aspect-based Sentiment Classification (MASC) and Joint Multimodal Aspect-Sentiment Analysis (JMASA). We illustrate the definitions of all the sub-tasks with a specific example in Figure~\ref{fig:absa}.

\textbf{Multimodal Aspect Term Extraction.}
As shown in Figure~\ref{fig:absa}, MATE aims to extract all the aspect terms mentioned in a sentence.
Given the multimodal input includes a n-words sentence $S = ( w_{1}, w_{2},…, w_{n})$ and a corresponding image $I$, the goal of MATE is to predict the label of each word in scheme $y_{i} \in \left\{B, I, O \right\}$, where $B$ indicates the beginning, $I$ indicates the inside and the end of an aspect term, $O$ means non-target words. 

Inspired by text-based aspect term extraction methods~\cite{142,143,144}, MATE approaches usually view this task as a sequence labeling problem. How to utilize visual information to improve the accuracy of aspect term recognition is the key to this task. Some studies~\cite{149,145} focused on named entity recognition suggest using ResNet encoding to leverage whole image information to enhance the representation of each word. Various neural network-based methods have been developed, including those using recurrent neural networks~\cite{145,146}, Transformers~\cite{96,147,148}, and graph neural networks~\cite{150}. Conditional Random Fields (CRF) are widely used in sequence labeling tasks because CRF considers the correlations between labels in neighborhoods. For example, an adjective has a greater probability of being followed by a noun than a verb in POS tagging task.
Using $Y=( y_{1}, y_{2},…, y_{n})$ represents a generic sequence of labels for input $S$. Given sequence $S$, all the possible label sequences $Y$ can be calculated by the following equation:
\begin{gather}
    p(Y|S) = \frac{\prod_{i=1}^{n}\Omega(y_{i-1},y_{i},X)}{\sum_{y^{'}\in Y}\prod_{i=1}^{n}\Omega(y^{'}_{i-1},y^{'}_{i},X)},
\end{gather}
where $\Omega(y_{i-1},y_{i},X)$ and $\Omega(y^{'}_{i-1},y^{'}_{i},X)$ are potential functions.

However, these methods are relatively independent and often focus more on entity information while neglecting the emotional information of the target. Therefore, as research progresses, more scholars in multimodal scenarios are not only extracting aspect terms but also jointly performing corresponding sentiment classification.

\begin{figure*}[h]
    \centering
    \includegraphics[width=8cm]{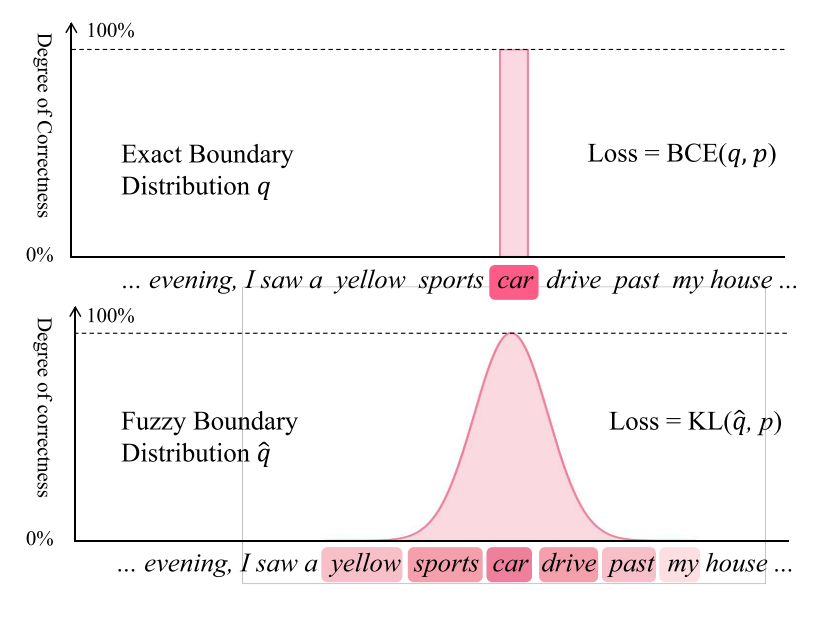}
    \caption{Illustration of exact boundary and fuzzy boundary.}
    \label{fig:mate_fuse}
\end{figure*}

\begin{figure*}[h]
    \centering
    \includegraphics[width=16cm]{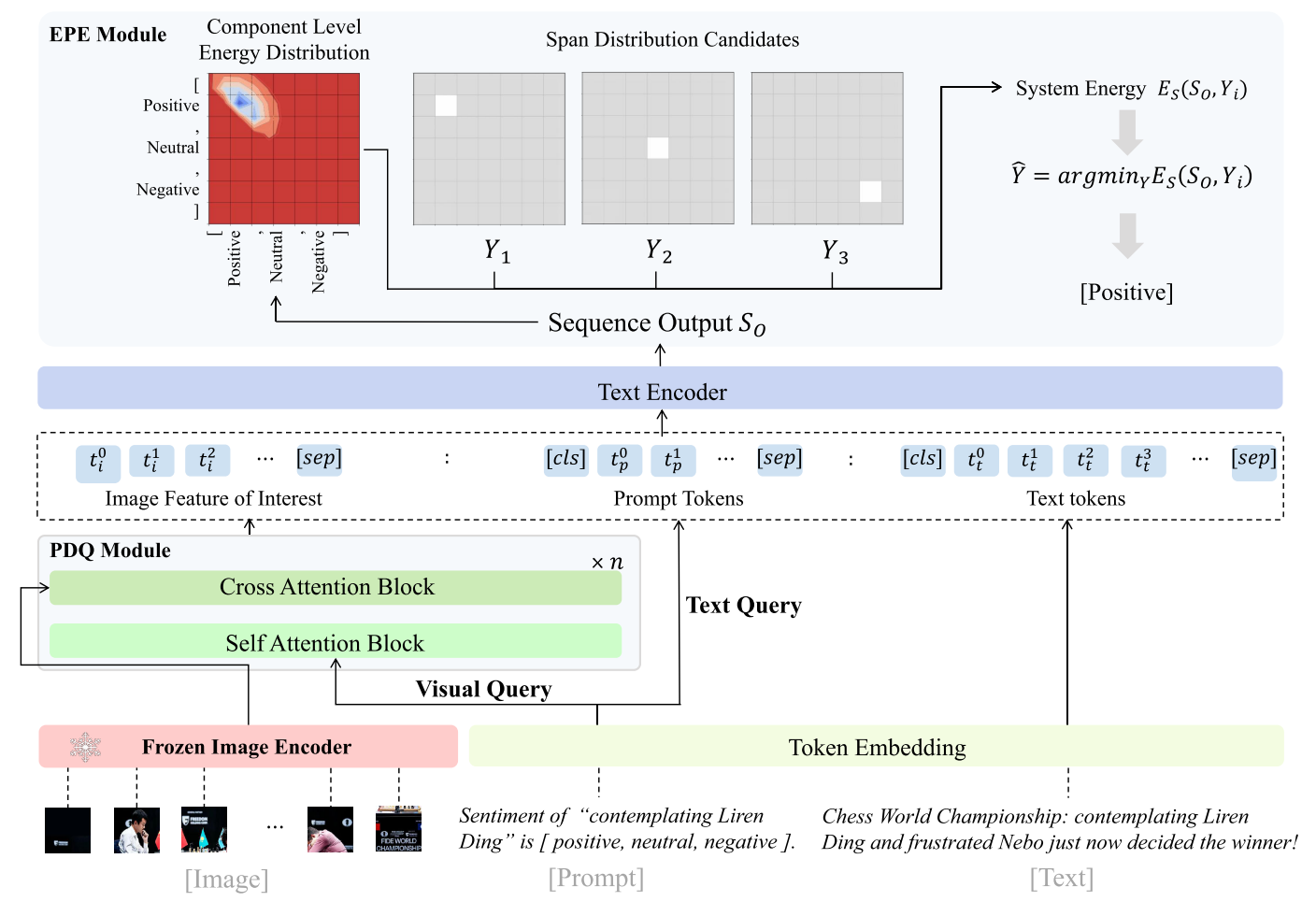}
    \caption{Overview of DQPSA model~\cite{151} architecture.}
    \label{fig:mate}
\end{figure*}

For example, inspired by the fuzzy span universal information extraction (FSUIE) framework~\cite{152} which focuses on the limited span length features in information extraction and proposes fuzzy span loss and fuzzy span attention, DQPSA~\cite{151} addresses MATE and MASC under a unified framework as span recognition, dispensing with the complex sequence generation structure. 
As shown in Figure~\ref{fig:mate}, To capture the semantical relevance of start and end boundaries within a span, DQPSA introduce the idea of Energy Based Model~\cite{216} to give better span scores and propose the Energy based Pairwise Expert that predicts span based on pairwise stability.

\textbf{Multimodal Aspect-based Sentiment Classification.}
As shown in Figure~\ref{fig:absa}, MASC aims to identify the sentiment polarity of a given aspect term in a sentence. Problem formalization as follows: Given a set of multimodal samples $S = \lbrace X_{1}, X_{2},..., X_{|S|}\rbrace$, where $|S|$ is the number of samples. And for each sample, we are given an image $V \in \mathbb{R}^{3 \times H \times W} $ where $3$, $H$ and $W$ represent the number of channels, height and width of the image, and an $N$-word textual content $T = (w_{1},w_{2},...,w_{N})$ which contains an $M$-word sub-sequence as target aspect $A = (w_{1},w_{2},..,w_{M})$. The goal of MASC is to learn a sentiment classifier to predict a sentiment label $y \in \lbrace Positive, Negative, Neutral\rbrace$ for each sample $X = (V, T, A)$.

Different from text-based aspect sentiment classification~\cite{153,154,155}, it is challenging to effectively discover visual sentiment information and fuse it with textual sentiment information. In \cite{77}, the authors constructed the Multi-ZOL dataset for the MASC task. This dataset collects and organizes comments about smartphones from the ZOL.com business portal website. At the same time, they proposed a multimodal interactive memory network(MIMN) based on an attention mechanism to capture the information interaction between different modalities. In addition, other researchers \cite{78,79} have proposed models like the LSTM-based ESAFN model and the Transformer-based TomBERT model for the MASC task, enhancing the interaction of inter-modal and intra-modal sentiment information is the core of these models.
The TomBERT model treat the hidden states of the target aspect $A$ as queries, and the regional image features $H^{V}$ as keys and values, such that the target is leveraged to guide the model to align it with the appropriate regions.
\begin{gather}
    ATT_{i}(A,H^{V}) = softmax(\frac{[W_{Q_{i}}A]^{T}[W_{K_{i}}H^{V}]}{\sqrt{d//m}})[W_{V_{i}}H^{V}]^{T}, 
\end{gather}
where ${W_{Q_{i}}, W_{K_{i}}, W_{V_{i}}} \in R^{d/m\times d}$ are parameters, $m$ represent the attention head number, $d$ is the input embeddings dimension. 

Compared with other multimodal tasks such as image and text retrieval, the sentiment annotation used in the MASC task lack strong supervision signals for cross-modal alignment. This issue makes it difficult for MASC models to learn cross-modal interactions and causes models to learn the bias brought by the image. In \cite{32}, the authors propose a new method to utilize visual modalities, the image caption generation module in their model undertakes the task of cross-modal alignment. They convert images into text descriptions based on the idea of cross-modal translation.
\begin{gather}
    C = Caption\_Transformer(V), 
\end{gather}
where $C$ and $Caption_transformer$ denote the output image captions and the transformer-based image caption generator, respectively.
In sentence-pair classification mode, input to pre-trained language model takes the sentence-pair form
\begin{gather}
    [CLS]w_{1}^{T} , w_{2}^{T} . . . w_{T.len}^{T} [SEP]w_{1}^{C} , w_{2}^{C} . . . w_{C.len}^{C} [PAD] . . . [PAD], 
\end{gather}
where $w_{i}^{T}$ are the tokens of the input text, and $w_{i}^{C}$ are the tokens of the image caption. In~\cite{98}, the authors continued with the idea of modal transformation and employed facial emotions as a supervised signal for learning visual emotions.

\begin{figure*}
    \centering
    \includegraphics[width=16cm]{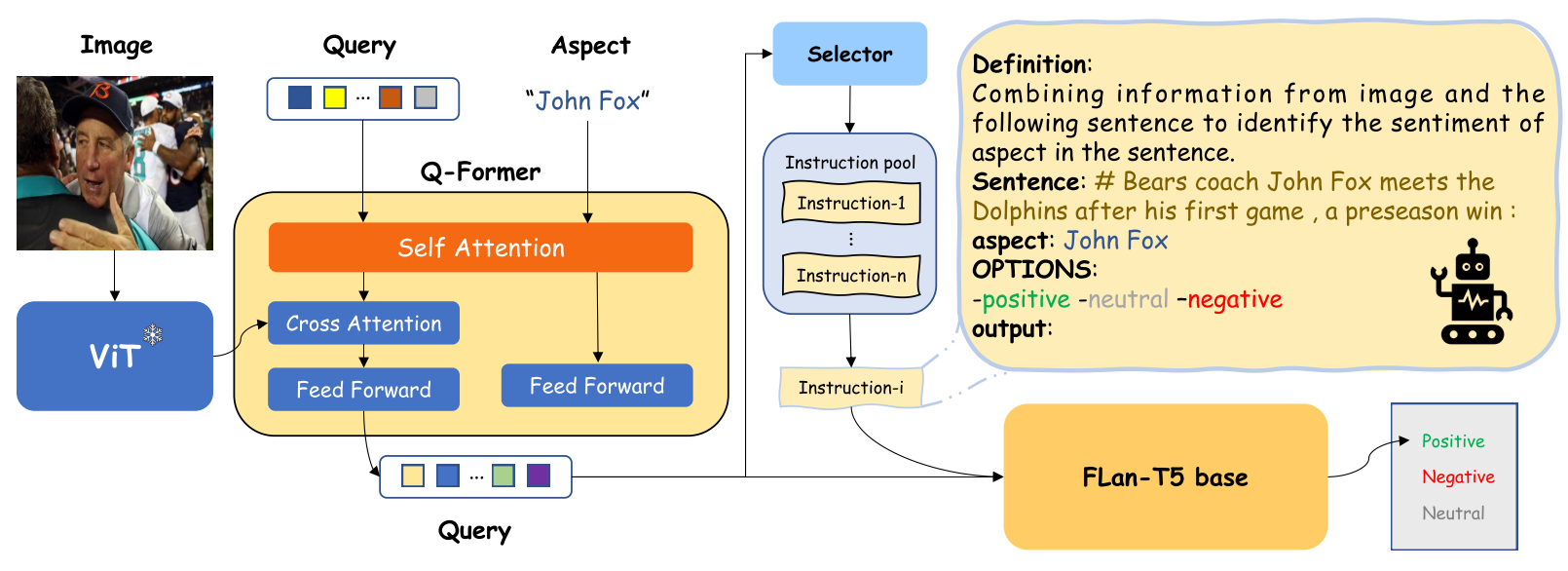}
    \caption{Overview of A$^{2}$II model~\cite{160} architecture for MASC task.}
    \label{fig:masc}
\end{figure*}

As large language models evolve, LLMs and LMMs have adapted to various tasks~\cite{156,157,158,159}, but their use in the MASC task is still in its initial stages. In~\cite{160}, the authors attempt to apply the instruction tuning paradigm to the MASC task and utilize the capabilities of LMMs to mitigate the limitations of text and image modality fusion. As shown in Figure~\ref{fig:masc}, to address the potential irrelevance between aspects and images, a plug-and-play selector is proposed to autonomously choose the most suitable instruction from the instruction pool, thereby reducing the impact of irrelevant image noise on the final sentiment classification outcome.

\textbf{Joint Multimodal Aspect-Sentiment Analysis.}
As shown in Figure~\ref{fig:absa}, JMASA aims to extract all aspect terms and their corresponding sentiment polarities simultaneously. Problem formalization as follows: Given a collection of multimodal sentence-image pairs, denoted as
$M$. Each pair $m_{i} \in M$ comprises a sentence $Si = ( w_{1}, w_{2},…, w_{n})$ and a corresponding image $v_{i}$. The objective of JMASA is to predict the corresponding aspect-sentiment pair $y =(y_{1}, y_{2},…, y_{n})$ for each sentence-image pair. Here, $y_{i} \in \left\{ B\text{-}POS, I\text{-}POS, B\text{-}NEG, I\text{-}NEG, B\text{-}NEU, I\text{-}NEU\right\} \cup {O}$. In this case, $B$ refers to the initial token of the aspect term, $I$ indicates tokens within the specific aspect term and $O$ indicates words ``outside'' the specific aspect. Moreover, $POS$, $NEU$, and $NEG$ are abbreviations for positive, neutral, and negative sentiments associated with the specific aspect.

As a pioneer, in~\cite{161}, the authors proposed joint multimodal aspect-sentiment analysis, which jointly performs multimodal aspect term extraction and multimodal aspect sentiment classification.
Since it is a joint task with aspect terms extraction and aspect sentiment classification, the authors calculate two different sets of loss simultaneously as follows:
\begin{gather}
    L = -\sum^{k}_{i=1}y_{i}^{s}\log p_{i}^{str} - \sum^{k}_{i=1}y_{i}^{e}\log p_{i}^{end} -\sum^{m}_{t=1}\sum^{\epsilon}_{i=1}y_{ti}^{p}\log p_{ti}^{p}
\end{gather}
where $y^{s}, y^{e}, y^{p}$ are one-hot labels indicating golden start, end positions, true sentiment polarity separately, and $a, m$ are the number of sentence tokens, aspects respectively.

Benefiting from the advancements in visual-language pre-train models, in~\cite{80}, the authors have designed multimodal sentiment pre-training tasks and develop a unified multimodal encoder-decoder architecture pre-training model for JMASA. In~\cite{162}, the authors utilize a Cross-Modal Multi-Task Transformer (CMMT) to derive sentiment-aware features for each modality and dynamically control the impact of visual information on textual content during inter-modal interaction. However, the innate semantic gap between visual and language modalities remains a huge challenge for the use of these methods, in~\cite{163}, the authors believe that the aesthetic attributes of images potentially convey a more profound emotional expression than basic image features and propesd Atlantis. Some scholars~\cite{164} have also noticed the impact of image-text pair quality, finding that many studies have overestimated the importance of images due to the presence of many noise images unrelated to text in the datasets. Drawing from the concept of curriculum learning, they proposed a Multi-grained Multi-curriculum Denoising Framework (M2DF), which achieves denoising by adjusting the order of the training data. AOM~\cite{165} is designed with an aspect-aware attention module that simultaneously selects text tokens and image blocks semantically related to the aspect to detect semantic and emotional information related to the aspect, thereby reducing noise introduced during the cross-modal alignment process. RNG~\cite{166} to simultaneously reduce multi-level modality noise and multigrained semantic gap, design three constraints: (i) Global Relevance Constraint (GR-Con) based on text-image similarity for instance-level noise reduction, (ii) Information Bottleneck Constraint (IB-Con) based on the Information Bottleneck (IB) principle for feature-level noise reduction, and (iii) Semantic Consistency Constraint (SC-Con) based on mutual information maximization in a contrastive learning way for multi-grained semantic gap reduction. To bridge the semantic gap between modal spaces and address the interference of irrelevant visual objects at different scales, in~\cite{167}, the authors proposed a Multi-level Text-Visual Alignment and Fusion Network (MTVAF).
\begin{figure*}
    \centering
    \includegraphics[width=16cm]{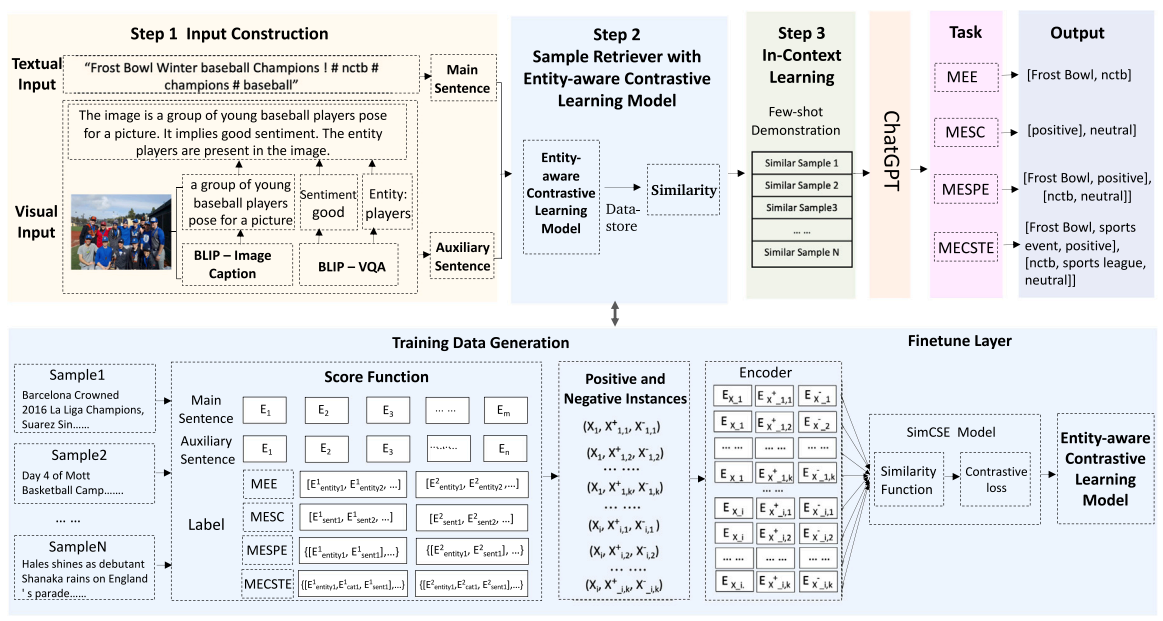}
    \caption{Overview of versatile ICL framework~\cite{171} architecture for JMASA task.}
    \label{fig:JmasC}
\end{figure*}

With the help of LLMs, the JMASA task has also seen further development in recent years.
In~\cite{170}, the authors found that compared to converting MASC into a masked language modeling (MLM) task based on limited sentiment categories with PVLM~\cite{169} and UP-MPF~\cite{168}, MLM is not suitable for JMASA and MATE tasks. They proposed a novel Generative Multimodal Prompt (GMP) model. In~\cite{171}, the authors explored the potential of using the representative large language model ChatGPT for In-Context Learning (ICL) on the JMASA task. As shown in Figure~\ref{fig:JmasC}, they developed a versatile ICL framework, incorporating zero-shot learning task instructions and expanded it to few-shot learning by adding some demonstration samples to the prompts. Additionally, to enhance the ICL framework's performance in few-shot learning scenarios, they further developed an entity-aware contrastive learning model to effectively retrieve demonstration samples similar to each test sample.

We summarize the commonly used datasets for fine-grained image-text sentiment analysis, including Multi-ZOL, Twitter-15 and Twitter-17:

\textbf{Multi-ZOL} collects and organizes comments about smartphones from the ZOL.com business portal website. Multi-ZOL dataset includes sentiment ratings for six aspects, such as price-performance ratio, performance configuration, battery life, appearance and feeling, photographing effect, and screen. For each aspect, the comment has an integer sentiment score
from 1 to 10, which is used as the sentiment label. 

\textbf{Twitter-2015} and \textbf{Twitter-2017} datasets are commonly used datasets for fine-grained image-text sentiment analysis tasks. These datasets are collected from English tweets on the social media platform Twitter and are in the form of image-text pairs. The datasets provide annotations for aspects mentioned in the text. Sentiment labels are categorized into three classes: positive, neutral and negative. Specifically, the Twitter-2015 dataset contains 5,338 tweets with images, while the Twitter-2017 dataset contains 5,972 tweets with images.

\subsection{Audio-Image-Text Sentiment Analysis}
Audio-image-text (video) sentiment analysis differs from image-text sentiment analysis in two main aspects: 1) Different data emphasis. Existing text-image sentiment datasets are drawn from social media and e-commerce platforms, covering a wide range of content. In contrast, visual information in video sentiment datasets often focuses on the facial expressions and body movements of speakers. 2) Videos can be considered as temporal sequences of text-image pairs, necessitating considerations of intra-modal emotional factors in audio sequences and video frame sequences, as well as alignment relationships between text, video frames, and audio over time. Video-based sentiment analysis primarily includes sentiment classification and emotion classification tasks. Sentiment classification involves three, five, or seven-category classification tasks, while emotion classification comprises multi-label emotion recognition (where each sample corresponds to multiple emotion labels) and single-label emotion recognition. Common emotion labels include happiness, surprise, and anger. Problem formalization as follows: 

In audio-image-text sentiment analysis tasks, the input is utterance consisting of three modalities: textual, acoustic and visual modality, where $m\in\{t,a,v\}$ . The sequences of these three modalities are represented as triplet $(T,A,V)$, including $T \in \mathbb{R}^{N_t\times d_t}$, $A \in \mathbb{R}^{N_a\times d_a}$ and $V \in \mathbb{R}^{N_v\times d_v}$ where $N_m$ denotes the sequence length of corresponding modality and $d_m$ denotes the dimensionality. The goal of audio-image-text sentiment analysis tasks is to learn a mapping $f(T,A,V)$ to infer the sentiment score $\hat{y}\in \mathbb{R}$.

As the audio-image-text sentiment analysis methods proposed by scholars in recent years generally cater to both sentiment classification and emotion classification tasks, this paper will review the existing multimodal sentiment analysis methods around two core themes: cross-modal sentiment semantic alignment and multimodal sentiment semantic fusion.

\subsubsection{Cross-modal Sentiment Semantic Alignment}
Cross-modal sentiment semantic alignment methods aim to explore the associations between emotional information across different modalities, analyze the corresponding relationships between them (alignment relationship modeling), and reduce the semantic distance between representations across modalities (semantic representation alignment). Cross-modal sentiment semantic alignment methods can help overcome the challenges brought by the semantic gap of heterogeneous modalities and are a prerequisite for multimodal sentiment semantic fusion methods. Specifically, by exploring the alignment relationships between different modal sentient semantic representations, these methods can help the fusion model ignore irrelevant information and focus on modeling effective information. By bringing emotional semantic representations closer in the representation space, these methods can reduce modal differences between representations, lower the difficulty of fusion, and increase fusion efficiency. This paper surveys existing cross-modal  sentiment semantic alignment methods and categorizes them into three types based on different alignment strategies and purposes: attention-based alignment, contrastive learning-based alignment, and cross-domain transfer learning-based alignment.

\textbf{Attention-based alignment.}
The attention mechanism has been proven to be an effective method for cross-modal semantic alignment in the field of multimodal learning~\cite{179}. Not only can the attention mechanism learn to adapt the alignment relationships for specific tasks through the optimization of task-specific objective functions, but it can also provide a degree of interpretability by outputting attention weights. For example, in the field of image captioning, the attention mechanism focuses on relevant areas when generating text words~\cite{180}, demonstrating the alignment relationship between words and image regions. Inspired by related research in the multimodal learning field, in~\cite{43}, the authors proposed using a cross-modal attention mechanism to learn the alignment relationships between pairs of modalities and developed a transformer-based multimodal sentiment analysis model named MulT. The core of the MulT model lies in modeling cross-modal alignment relationships by inserting cross-modal attention layers into the transformer module, allowing dynamic alignment and fusion of fine-grained sentiment information from various modalities. First, use $\alpha$ and $\beta$ to represent two different modalities, define the $\mathrm{Query}$s as $Q_\alpha = X_\alpha W_{Q_\alpha}$, $\mathrm{Key}$s as $K_\beta = X_\beta W_{K_\beta}$, and $\mathrm{Value}$s as $V_\beta = X_\beta W_{V_\beta}$, where $W_{Q_\alpha} \in \mathbb{R}^{d_\alpha \times d_k}, W_{K_\beta} \in \mathbb{R}^{d_\beta \times d_k}$ and $W_{V_\beta} \in \mathbb{R}^{d_\beta \times d_v}$ are weights. The latent adaptation from $\beta$ to $\alpha$ is presented as the crossmodal attention $Y_\alpha := \text{CM}_{\beta \rightarrow \alpha}(X_\alpha, X_B) \in \mathbb{R}^{T_\alpha \times d_v}$:
\begin{gather}
      Y_\alpha = \text{CM}_{\beta \rightarrow \alpha}(X_\alpha, X_\beta)= \text{softmax}\left(\frac{Q_\alpha K_\beta^\top}{\sqrt{d_k}}\right) V_\beta = \text{softmax}\left(\frac{X_\alpha W_{Q_\alpha} W_{K_\beta}^\top X_\beta^\top}{\sqrt{d_k}}\right) X_\beta W_{V_\beta}.
\end{gather}
Note that $Y_\alpha$ has the same length as $Q_\alpha$, but is meanwhile represented in the feature space of $V_\beta$. Specifically, the scaled (by $\sqrt{d_k}$) softmax computes a score matrix $\mathrm{softmax\,}(
\cdot) \in \mathbb{R}^{T_\alpha \times T_\beta}$, whose $(i,j)$-th entry measures the attention given by the $i$-th time step of modality $\alpha$ to the $j$-th time step of modality $\beta$. Hence, the $i$-th time step of $Y_\alpha$ is a weighted summary of $V_\beta$, with the weight determined by $i$-th row in $\mathrm{softmax}(\cdot)$.

Building on the cross-modal attention mechanism designed in MulT, in~\cite{181}, the authors introduced the cubic attention mechanism, which generates a three-dimensional attention tensor through parameter computations, representing the alignment information among the three modal representations.

\textbf{Contrastive learning-based alignment.} 
Contrastive learning achieves cross-modal representation alignment by bringing the representations of positive examples closer together and pushing the representations of negative examples farther apart. A classic model in the field of multimodal learning, CLIP~\cite{116}, uses contrastive learning to align the semantic representations of text and image modalities, significantly enhancing the quality of image representations and achieving excellent results in tasks such as zero-shot image classification. Inspired by this, the field of audio-image-text sentiment analysis has adopted contrastive learning methods for sentiment semantic representation alignment. In~\cite{182}, the authors proposed achieving cross-modal emotional semantic alignment by bringing closer the representations of different modalities within the same sample. In~\cite{183}, the authors suggested using the text-audio and text-image modal information of input samples to predict the corresponding image and audio representations of the samples, then aligning the predicted representations with the actual ones and distancing representations from different samples, thereby aligning the semantic representations of different modalities within the same sample. The proposed model contains two key modules, the uni-modal coding drive the model to focus on
informative features which then implicitly filter out inherent noise and produces robust and effective uni-modal representation for acoustic and visual modalities. 

Given a batch set $F_{uni}=\{F_u^0, F_u^1, ..., F_u^{n-1}\}$, noted that there is a single positive key $F^\textit{\dag}_u$ (as $k^+$) that each encoded query $F_u^i$ (as $q$, $i\in[1,n]$) matches, while the other representations $F_u^j$ ($j\in[0,n]$ and $j\neq i$) in the same batch are considered as negative key samples $k^-$. With the similarity measured by dot product, the uni-modal instance contrastive loss $\mathcal L_{uni}$ in:
\begin{gather}
    \mathcal L^u_{uni} \triangleq -\log \frac{\exp(q\cdot k^+/\tau)}{\sum_{i=1}^{n}\exp(q\cdot q^i/\tau)} = -\mathop{\mathbb{E}}\limits_{F_{uni}}\left[ \log \frac{\exp(F_u\cdot F^\textit{\dag}_u/\tau)}{\sum_{i=1}^{n}\exp(F_u\cdot F_u^i/\tau)}\right]
\end{gather}
where $\tau$ is a temperature hyper-parameter that controls the probability distribution over distinct instances. Due to $u\in\{a,v\}$, the final uni-modal instance contrastive loss $\mathcal L_{uni}=\mathcal L^a_{uni}+\mathcal L^v_{uni}$.

The cross-modal prediction captures commonalities among different modalities and outputs predictive representation full of interaction dynamics. Each query $q$ has a corresponding key as $k^+$ while the other representations in the same batch are seen as $k^-$. Similar with uni-modal instance contrastive loss $\mathcal L_{uni}$, the cross-modal instance contrastive loss $\mathcal L_{cross}$ is presented as:

\begin{gather}
    \mathcal L_{cross} \triangleq -\mathop{\mathbb{E}}\limits_{F_{cross}}\left[ \log \frac{\exp(F_{c}\cdot F^+_{c}/\tau)}{\sum_{i=1}^{n}\exp(F_c\cdot F_c^{i}/\tau)}\right]
\end{gather}
where $F_c\backslash F_c^+\in\{P_{u},G_{u}\}$, $P_{u}$ represent the prediction while $G_{u}$ represent the target, $u\in\{a,v\}$ and $F_{cross}=\{F_c^1, ..., F_c^n\}$.

\textbf{Cross-domain transfer learning-based alignment.}
The field of cross-domain transfer learning primarily studies how to align the sample spaces of target domains with those of source domains so that classifiers trained in the source domains can be directly reused in the target domains. The objectives of this field align broadly with those of cross-modal sentiment representation alignment, hence some studies have explored using cross-domain transfer learning methods for sentiment semantic representation alignment. In~\cite{184}, considering the rich information content of textual representations,  the authors proposed using Deep Canonical Correlation Analysis (DCCA) to align audio and visual representations with textual representations, thereby enhancing the audio and visual representations. In~\cite{185}, the authors explored using a metric-based domain transfer method, utilizing Central Moment Discrepancy (CMD) to design a loss function that aligns the representations of the three modalities within the same sample. The overall learning of the model is performed by minimizing:
\begin{gather}
        \mathcal{L} =  \mathcal{L}_{\text{task}} + \alpha \,  \mathcal{L}_{\text{sim}}  + \beta \, \mathcal{L}_{\text{diff}} + \gamma\, \mathcal{L}_{\text{recon}} 
\end{gather}
Here, $\alpha, \beta, \gamma$ are the interaction weights that determine the contribution of each regularization component to the overall loss $\mathcal{L}$. 
Minimizing the \textit{similarity loss} $\mathcal{L}_{\text{sim}}$ reduces the discrepancy between the shared representations of each modality. This helps the common cross-modal features to be aligned together in the shared subspace.
Difference Loss $\mathcal{L}_{\text{diff}}$ is to ensure that the modality-invariant and -specific representations capture different aspects of the input. 
As the difference loss is enforced, there remains a risk of learning trivial representations by the modality-specific encoders. To avoid this situation, the authors add a reconstruction loss $\mathcal{L}_{\text{recon}}$ that ensures the hidden representations to capture details of their respective modality.
The task-specific loss $\mathcal{L}_{\text{task}}$ estimates the quality of prediction during training.

In~\cite{186} the authors have employed adversarial learning methods to align sentiment semantic representations across different modalities.

\subsubsection{Multimodal Sentiment Semantic Fusion}
Multimodal sentiment semantic fusion aims to efficiently aggregate sentiment information from different modalities to achieve comprehensive and accurate sentiment understanding. The challenge of fusion lies in how to fully capture the complex interactions among multimodal sentiment semantic information, thereby facilitating sentiment reasoning and prediction. This paper surveys existing multimodal sentiment semantic fusion methods and categorizes them into three types: tensor-based fusion, fine-grained temporal interaction modeling fusion, and pre-trained model-based fusion.

\textbf{Tensor-based fusion.}
In the early stages of audio-image-text sentiment analysis research, considering the small scale of datasets and limited computational resources, researchers represented the raw inputs of each modality as a single emotional semantic representation before proceeding to multimodal emotional semantic representation fusion. The simplest fusion strategy was to directly concatenate the emotional semantic representations of different modalities, but this method did not explicitly model the higher-order interactions between emotional information from different modalities. To address this issue, in~\cite{33}, the authors proposed using the outer product of vectors to fuse different modal representations, thereby modeling interactions among unimodal, bimodal, and trimodal emotional semantic representations simultaneously. However, this method, due to the complexity of the outer product operation being tied to the product of input vector dimensions, resulted in high computational costs and slow efficiency. Subsequent work has made efficiency improvements. In~\cite{187}, the authors proposed the LMF fusion method, which accelerates the fusion process of multimodal emotional representations through low-rank decomposition. 
\begin{figure*}[h]
    \centering
    \includegraphics[width=16cm]{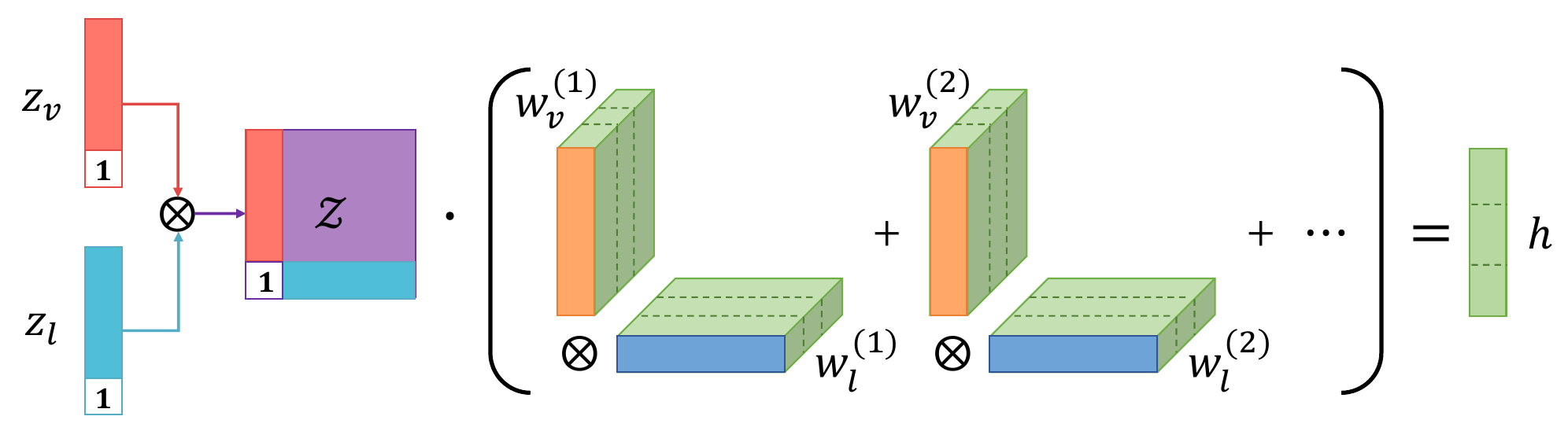}
    \caption{Decomposing weight tensor into low-rank factors~\cite{187}.}
    \label{fig:tfusion}
\end{figure*}

In~\cite{188}, the authors introduced a three-stage multimodal emotional representation fusion strategy consisting of representation slice grouping, intra-group representation slice fusion, and global representation fusion. Representation slice grouping involves splitting the representations of each modality into the same fixed number of small groups, numbering them, and then locally fusing representation slices of the same number from different modalities together. This approach reduces the dimensions of representations to be fused later, thereby enhancing fusion efficiency. Intra-group representation slice fusion uses the outer product method to fuse the representation slices of the three modalities within the group, which, due to the smaller feature dimensions, significantly speeds up the fusion process. Finally, Long Short-Term Memory (LSTM) networks are used to perform global representation fusion of the different groups after fusion. This method reduces the computational complexity of the tensor outer product fusion method to some extent through block processing.

\textbf{Fine-grained temporal interaction modeling fusion.}
This type of fusion method focuses on capturing more localized, fine-grained interactions of multimodal information. These methods first obtain fine-grained representations corresponding to each time step of each modality, and then perform multimodal sentiment semantic fusion based on these representations to capture the interactions between cross-modal and cross-temporal sentiment information. In~\cite{34}, the RAVEN model is a typical method in this series of research. The authors found that the same words can convey different emotional messages when accompanied by different tones or expressions. Driven by this motivation, they designed a network that improves the word representations by dynamically integrating the fine-grained representations of visual and auditory modalities into each word vector through a cross-modal gating mechanism, thereby achieving the goal of infusing non-verbal emotional information into word representations. For a word $\mathbf{L}^{(i)}$, the nonverbal shift vector $\mathbf{h}^{(i)}_m$ is calculated as follows:
\begin{gather}
        \mathbf{h}^{(i)}_m = w_v^{(i)} \cdot (\mathbf{W}_{v} \mathbf{h}^{(i)}_v) + w^{(i)}_a \cdot (\mathbf{W}_{a} \mathbf{h}^{(i)}_a) + \mathbf{b}^{(i)}_h
\end{gather}
where $\mathbf{W}_{v}$ and $\mathbf{W}_{a}$ are weight matrices for the visual and acoustic embedding and $\mathbf{b}^{(i)}_h$ is the bias vector.

In\cite{189}, considering that audio and visual inputs might contain noise at certain time steps, like background noise in speech, the authors proposed a reinforcement learning-based gating unit to control the information fusion between fine-grained representations of different modalities. The gating mechanism allows for dynamic sentiment representation fusion by controlling whether the representation of the current word incorporates information from a particular modality. Unlike the previous two works, which focus on capturing interactions of multimodal fine-grained sentiment representations associated with individual words, in~\cite{190}, the authors model the feature interactions between multimodal fine-grained sentiment representations of multiple consecutive words within a window and use a memory neural network to model global information.

\textbf{Pre-trained model-based fusion.}
Pre-trained language models have demonstrated strong language understanding capabilities, and researchers believe they also hold great potential for multimodal language understanding. To explore the capabilities of pre-trained language models in the field of multimodal sentiment analysis, in~\cite{191} the authors, inspired by the RAVEN method, designed a gating mechanism for pre-trained language models. The aim is to inject multimodal information into the intermediate layer word representations of the pre-trained language models to fully leverage their strong language modeling capabilities for efficient multimodal emotional understanding. In~\cite{192}, the authors proposed a cross-modal efficient attention mechanism that uses the output representations of pre-trained language models to compress the input sequences of visual and audio features, thereby enhancing the model's computational efficiency.

\begin{figure*}[h]
    \centering
    \includegraphics[width=16cm]{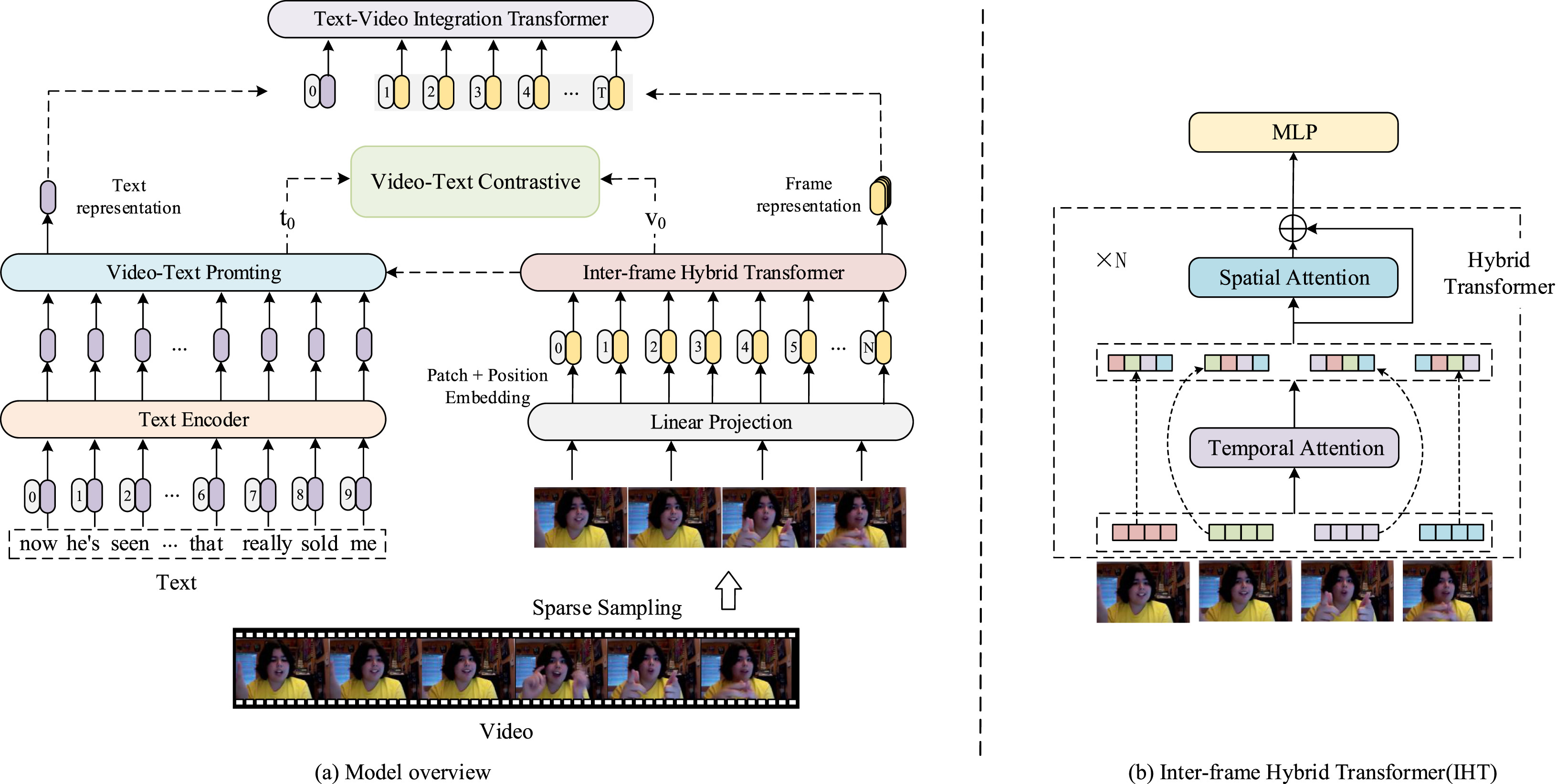}
    \caption{Overview of the VLP2MSA~\cite{204} architecture(a) and the Inter-frame Hybrid Transformer module(b). The model receives the original text and video frames as input and consists of a text encoder, a video encoder, a prompt module, a video-text contrastive, and a multimodal integration encoder. we propose an adapted and extended video encoder and a new fusion scheme that prompts and aligns before interacting to reduce the influence of modality heterogeneity on the fusion results.}
    \label{fig:vmsa}
\end{figure*}

To effectively extend LLMs and LMMs to multimodal sentiment analysis tasks and address two pressing challenges in the field: (1) the low contribution rate of the visual modality and (2) the design of an effective multimodal fusion architecture, scholars~\cite{204} have proposed an inter-frame hybrid transformer. As shown in Figure~\ref{fig:vmsa}, this transformer extracts spatiotemporal features from sparsely sampled video frames, focusing not only on facial expressions but also capturing body movement information.

\subsubsection{Audio-image-text Sentiment Analysis Datasets}
We summarize the commonly used datasets for audio-image-text image-text sentiment analysis, including ICT-MMMO, IEMOCAP, CMU-MOSI, CMU-MOSEI, MELD, CH-SIMS, CH-SIMS 2, M3ED, MER2023, EMER, MER2024, CMU-MOSEAS and UR-FUNNY.

\textbf{ICT-MMMO} dataset is collected from the YouTube website and defines seven sentiment labels based on sentiment polarity and intensity: positive (strong), positive, positive (weak), neutral, negative (weak), negative, and negative (strong). In~\cite{193}, the authors first addressed the task of tri-modal sentiment analysis and demonstrated that it is a feasible task that can benefit from the combined use of image, audio, and text modalities. This dataset forms the basis of their research.

\textbf{IEMOCAP} dataset is a multimodal video dialogue dataset collected by the SAIL lab at the University of Southern California. It contains about 12 hours of multimodal data, including video, audio, facial motion capture, and transcribed text. The dataset was collected through dialogues by 5 professional male actors and 5 professional female actors in pairs, engaging in either improvised or scripted dialogues, with a focus on emotional expression. The dataset includes a total of 4,787 improvised dialogues and 5,255 scripted dialogues, with an average of 50 sentences per dialogue and an average duration of 4.5 seconds per sentence. Each sentence in the dialogue segments is annotated with specific emotional labels, divided into ten categories including anger, happiness, sadness, and neutral.

\textbf{CMU-MOSI} and \textbf{CMU-MOSEI} are two commonly used datasets in the multimodal sentiment analysis area. The data is sourced from video blogs (vlogs) on the online sharing platform YouTube. These datasets primarily focus on coarse-grained multimodal sentiment classification tasks. CMU-MOSI dataset comprises 2,199 video segments extracted from 93 distinct videos. The video content consists of English comments posted by individual speakers. There are 41 female and 48 male speakers, mostly between the ages of 20 and 30, coming from diverse backgrounds (Caucasian, Asian, etc.). The videos are annotated by five annotators from the Amazon Mechanical Turk platform, and the annotations are averaged. Annotations cover seven categories of emotional tendencies ranging from -3 to +3. The CMU-MOSEI dataset is larger than the CMU-MOSI dataset, containing 23,453 video segments from 1,000 different speakers across 250 topics, with a total duration of 65 hours. The dataset includes both emotion labels and sentiment labels. Emotion labels include happiness, sadness, anger, fear, disgust, and surprise, while sentiment labels include sentiment binary classification, five classification, and seven classification annotations.

\textbf{MELD} dataset originates from the classic TV series Friends. It comprises a total of 1,443 dialogues and 13,708 utterances, with an average of 9.5 sentences per dialogue and an average duration of 3.6 seconds per sentence. Each sentence in the dialogue segments is annotated with one of seven emotional labels, including anger, disgust, sadness, happiness, neutral, surprise, and fear. Additionally, each sentence is also assigned a sentiment label, categorized as positive, negative, or neutral.

\textbf{CH-SIMS} dataset is a Chinese multimodal sentiment classification dataset with the unique feature of having both unimodal and multimodal sentiment labels. It consists of 60 original videos collected from movie clips, TV series, and various performance shows. These videos were clipped at the frame level to obtain 2,281 video segments. Annotators labeled each video segment for four modalities: text, audio, silent video, and multimodal. To avoid cross-modal interference during annotation, annotators could only access information from the current modality. They first performed unimodal labeling, followed by multimodal labeling. Although the dataset provides labels for each modal, its primary purpose is coarse-grained multimodal sentiment classification. The \textbf{CH-SIMS 2} dataset expands the CH-SIMS dataset. This dataset is larger in scale and more difficult, requiring the model to accurately integrate information from different modalities to predict the correct answer.

\textbf{M3ED} dataset includes 990 Chinese dialogue videos, totaling 24,449 sentences. Each sentence is annotated for six basic emotions (happiness, surprise, sadness, disgust, anger, and fear), as well as neutral emotion.

\textbf{MER2023} includes four subsets: Train \& Val, MERMULTi, MER-NOISE, and MER-SEMI. In the last subset, besides the labeled samples, it also contains a large amount of unlabeled data. The dataset annotates sentiment labels on each sample and focuses on challenges such as multi-label learning, noise robustness, and semi-supervised learning. Furthermore, they built upon MER2023 to create the \textbf{EMER} dataset, which not only annotates sentiment labels on each sample but also the reasoning process behind the labels. In \textbf{MER2024}, they expanded the dataset size and included a subset with multi-label annotations, attempting to describe the emotional states of characters as accurately as possible.

\textbf{CMU-MOSEAS} is the first large-scale multimodal language
dataset for Spanish, Portuguese, German and French, with 40, 000 total labelled sentences. It covers a diverse set topics and speakers, and carries supervision of 20 labels including sentiment (and subjectivity), emotions, and attributes. 

\textbf{UR-FUNNY} dataset is tailored for humor detection tasks, which are closely related to multimodal sentiment analysis. The dataset was collected from the TED website, selecting 8,257 humorous snippets from 1,866 videos and their transcribed texts, and additionally, 8,257 non-humorous segments were randomly chosen. The total duration of the dataset is 90.23 hours, encompassing 1,741 different speakers and 417 distinct topics.

\subsection{Multimodal Sarcasm Detection}

Sarcasm detection task initially only focused on the textual context~\cite{49,50,51}, with scholars noting that common ironic sentences often juxtapose positive phrases with negative contexts. For example, in the sentence ``I'm so happy I'm late for work'', the presence of the positive phrase ``happy'' within the negative context of being late for work makes it easily recognizable as sarcasm. In most cases, the sentiment signals conveyed by different modalities in multimodal data are consistent. However, there are instances of inconsistency, necessitating sentiment disambiguation across modalities. Multimodal sentiment disambiguation is essentially a classification task. Multimodal sentiment inconsistency can be categorized into two types: complete sentiment conflict, defined as multimodal irony recognition tasks, and instances where some modalities convey 'neutral' sentiment polarities while others convey positive or negative sentiment polarities, which are typical cases of implicit sentiment expression. The multimodal sarcasm detection task formalization as follows: 

Multimodal sarcasm detection aims to identify if a given text associated with an image has sarcastic meaning. Formally, given a set of multimodal samples $D$, for each sample $d \in D$, it contains a sentence $T$ with $n$ words $\left\{ t_{1}, t_{2}, t_{3},..., t_{n}\right\}$ and an associated image $I$. The goal of model is to learn a multimodal sarcasm detection classifier to correctly predict the results of unseen samples.

In~\cite{53}, the authors introduced a multimodal sarcasm detection task for videos and compiled a corresponding dataset from television series. Considering the correlation between sentiment classification and sarcasm detection, in~\cite{54}, the authors proposed a multi-task framework to simultaneously recognize sarcasm and classify sentiment polarity. In~\cite{57}, the authors suggested identifying sarcasm by capturing incongruent emotional semantic cues across modalities, such as rolling one’s eyes while uttering praise. 

Additionally, some researchers have studied sarcasm in text and images; for example, in~\cite{55}, the authors introduced a multimodal sarcasm detection task for text and images and designed a multi-level fusion network to detect sarcasm. In~\cite{194}, the authors proposed the multimodal sarcasm detection model using two different computational frameworks based on SVM and CNN that integrate text and visual modalities. Identifying inconsistencies between modalities is key to multimodal sarcasm detection, and recent models can be categorized into those based on attention mechanisms and those using Graph Neural Networks (GNN). For example, in~\cite{195}, the authors introduced a BERT-based model with a cross-modal attention mechanism and a text-oriented co-attention mechanism to capture inconsistencies within and between modalities. In~\cite{196}, the authors designed a 2D internal attention mechanism based on BERT and ResNet to extract relationships between words and images. In~\cite{197}, the authors proposed a Transformer-based architecture to fuse textual and visual information. In terms of GNN, scholars in~\cite{198} built heterogeneous intramodal and cross-modal graphs (InCrossMG) for each multimodal example to determine the emotional inconsistencies within specific modalities and between different modalities, and introduced an interactive graph convolutional network structure to learn the relationships of inconsistencies in a joint and interactive manner within modal and cross-modal graphs. In~\cite{199}, the authors constructed heterogeneous graphs containing fine-grained object information of images for each instance and designed a cross-modal graph convolutional network. 
\begin{figure*}[h]
    \centering
    \includegraphics[width=8cm]{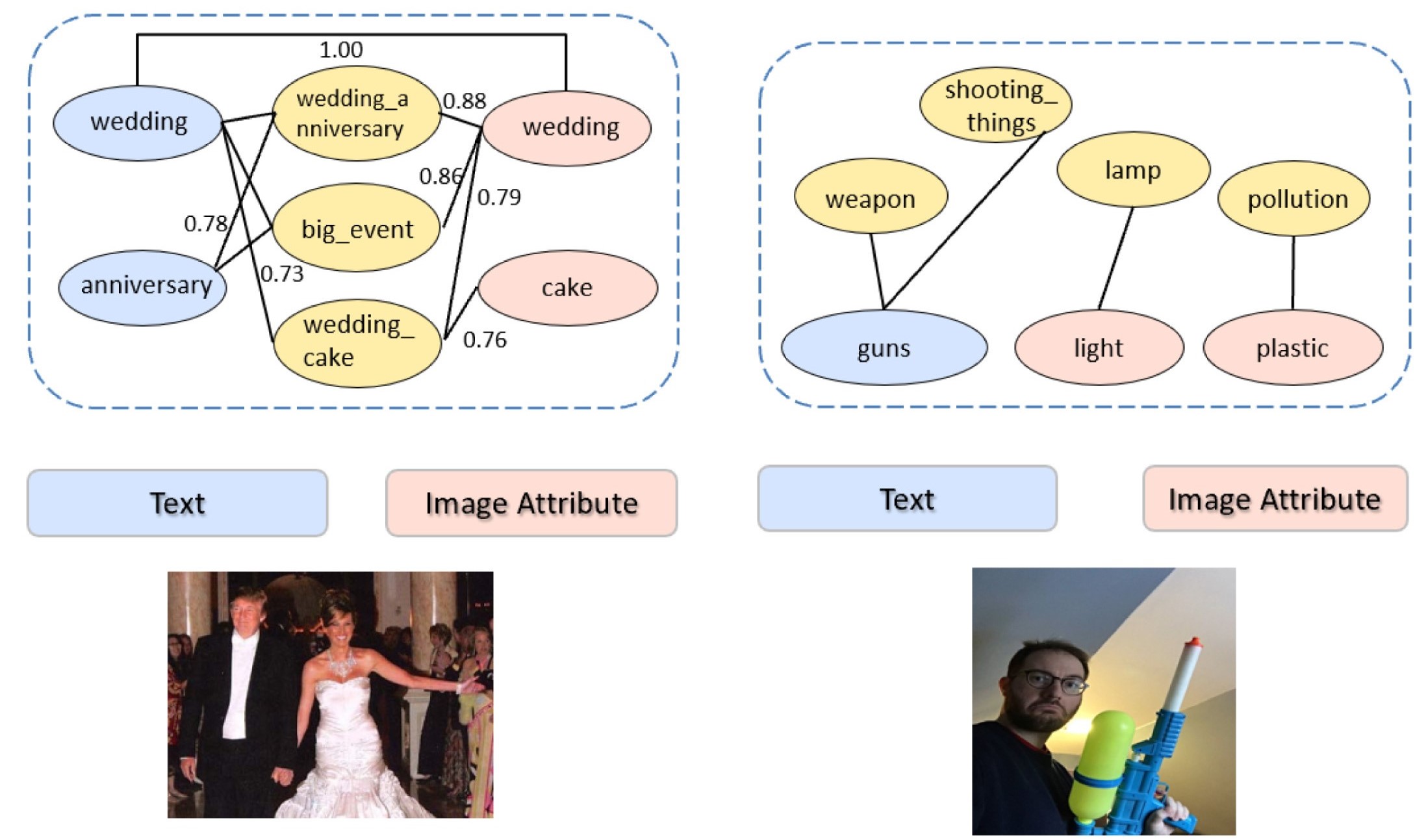}
    \caption{Different processing procedures for positive(sarcasm) and negative(non-sarcasm) samples in knowledge-based word-level cross-modal semantic similarity detection module~\cite{200}.}
    \label{fig:concept}
\end{figure*}

Additionally, some scholars~\cite{200} have proposed incorporating prior knowledge into multimodal sarcasm detection. They introduced KnowleNet, which utilizes the ConceptNet~\cite{201} knowledge base to integrate prior knowledge, and determines the relevance between images and text through sample-level and word-level cross-modal semantic similarity detection. They also incorporated contrastive learning to improve the spatial distribution of sarcastic (positive) and non-sarcastic (negative) samples. In~\cite{202}, the authors propose a lightweight multimodal interaction model with knowledge enhancement based on deep learning. In~\cite{203}, the authors proposed to introduce emotional knowledge into sarcasm detection and used sentiment dictionaries to obtain the sentiment vectors by evaluating the words extracted from various modalities, and then combined them with each modality. In~\cite{205}, the authors propsed multi-view CLIP that is capable of leveraging multi-grained cues from multiple perspectives (i.e., text, image, and textimage interaction view) for multi-modal sarcasm detection.
\begin{figure*}
    \centering
    \includegraphics[width=16cm]{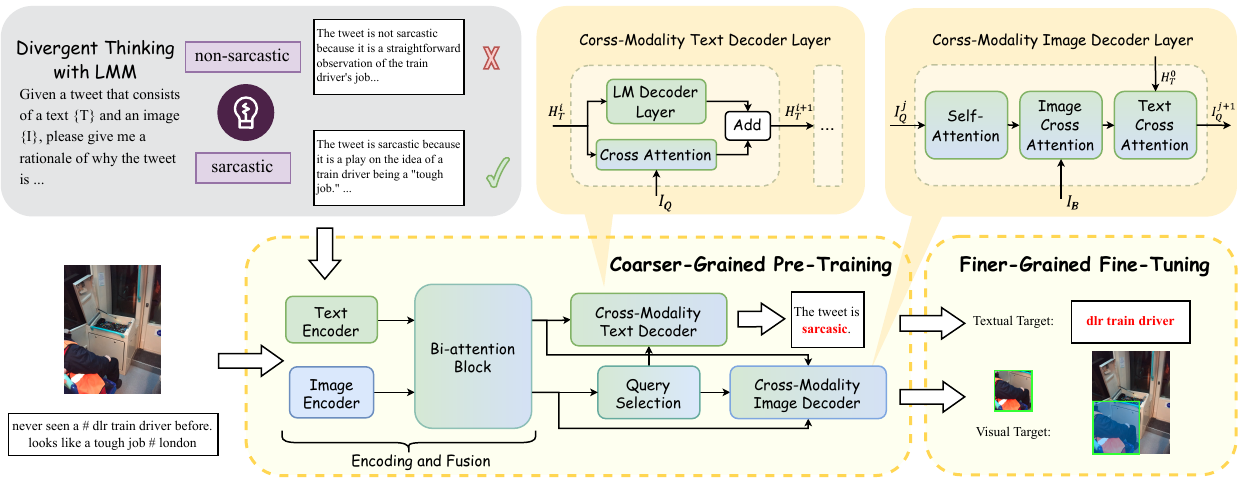}
    \caption{An overview of CofiPara for multimodal sarcasm target identification~\cite{207}.}
    \label{fig:sdllm}
\end{figure*}

In~\cite{206}, the authors tested the performance of some existing open-source LLMs and LMMs in the multimodal sarcasm detection task and proposed a generative multi-media sarcasm model consisting of a designed instruction template and a demonstration retrieval module based on the large language model. In~\cite{207}, the authors proposed
a versatile framework with a coarse-to-fine paradigm, by augmenting sarcasm explainability with reasoning and pre-training knowledge. 

We summarize the commonly used datasets for multimodal sarcasm detection, including MMSD, MMSD2.0 and MUStARD.

\textbf{MMSD} dataset is collected from the Twitter platform by searching for tweets in English that contain special tags indicating sarcasm, such as \#sarcasm, \#sarcastic, \#irony, \#ironic, to gather sarcastically labeled data, and collecting other tweets without these tags as non-sarcastic data. The dataset is annotated with a binary classification of ``sarcastic/non-sarcastic''. \textbf{MMSD2.0} fixed the shortcomings of MMSD, by removing the spurious cues and re-annotating the unreasonable samples. 

\textbf{MUStARD} is a multimodal sarcasm detection dataset primarily sourced from English sitcoms, including Friends, The Big Bang Theory, The Golden Girls, and non-sarcastic video content from the MELD dataset. The authors collected a total of 6,365 video clips from these sources and annotated them, including 345 sarcastic video clips. To balance the categories, an equal number of 345 non-sarcastic video clips were selected from the remaining clips, resulting in a dataset comprising 690 video segments. The annotations include the dialogue, speaker, context dialogue and its speaker, the source TV show, and a label indicating whether it is sarcastic. The rich annotation allows researchers to conduct a variety of learning tasks, including studying the impact of context and speakers on the task of sarcasm detection.

\begin{table}[]
\caption{Some text-centric multimodal sentiment analysis methods that have utilized LLMs.}
\label{methods}
\begin{tabular}{p{2cm}p{3.8cm}p{3.8cm}p{3.8cm}}
\hline
Method & Usage of LLMs & Advantage & Disadvantage \\ \hline
WisdoM~\cite{141}       & Zero-shot Learning: 1) Using ChatGPT to provide prompt templates. 2) Prompting LMMs to generate context using the prompt templates with image and sentence.	& Leverage the contextual world knowledge induced from the LMMs for enhanced Image-text Sentiment Classification.           & Due to hallucinations in LLMs, the contextual knowledge supplemented by LLMs and LMMs as knowledge sources may not be accurate. The adaptive incorporation of context requires further exploration.             \\ \hline
ChatGPT-ICL~\cite{171}       & Zero-shot Learning and Few-shot Learning: Using ChatGPT to predict final sentiment labels.              & This work explores the potential of ICL with ChatGPT for Multimodal Aspect-based sentiment analysis, achieves competitive performance while utilizing a significantly smaller sample size.          & The ICL framework exhibits a relatively limited capability for aspect term extraction tasks when compared to fine-tuned methods.             \\ \hline
A2II~\cite{160} & Full-Parameter Tuning: leverage the ability of LMMs to alleviate the limitation of cross-modal fusion              & This work explored an instruction tuning modeling approach for multimodal aspect-based sentiment classification task, and achieved impressive performance.          & The visual features extracted by the Q-Former structure, which queries based on aspect, may be mismatched, leading to the neglect of some visual emotional signals.             \\ \hline
CofiPara~\cite{207} & Zero-shot Learning: Using potential sarcastic labels as prompts to cultivate divergent thinking in LMMs, eliciting the relevant knowledge in LMMs for judging irony.              & Note the negative impact of the inevitable noise in LMMs, and use competitive principles to align the sarcastic content generated by LMMs with their original multimodal features to reduce the noise impact. View LMMs as modal converters, transforming visual information into text to help cross-modal alignment.          &     Viewing LMMs as a knowledge source largely depends on the capabilities of the LMMs themselves. Although effective measures have been taken to reduce the impact of noise from LMMs, a certain proportion of erroneous judgments are still caused by LMMs.         \\ \hline
\end{tabular}
\end{table}

\begin{figure*}
    \centering
    \includegraphics[width=16cm]{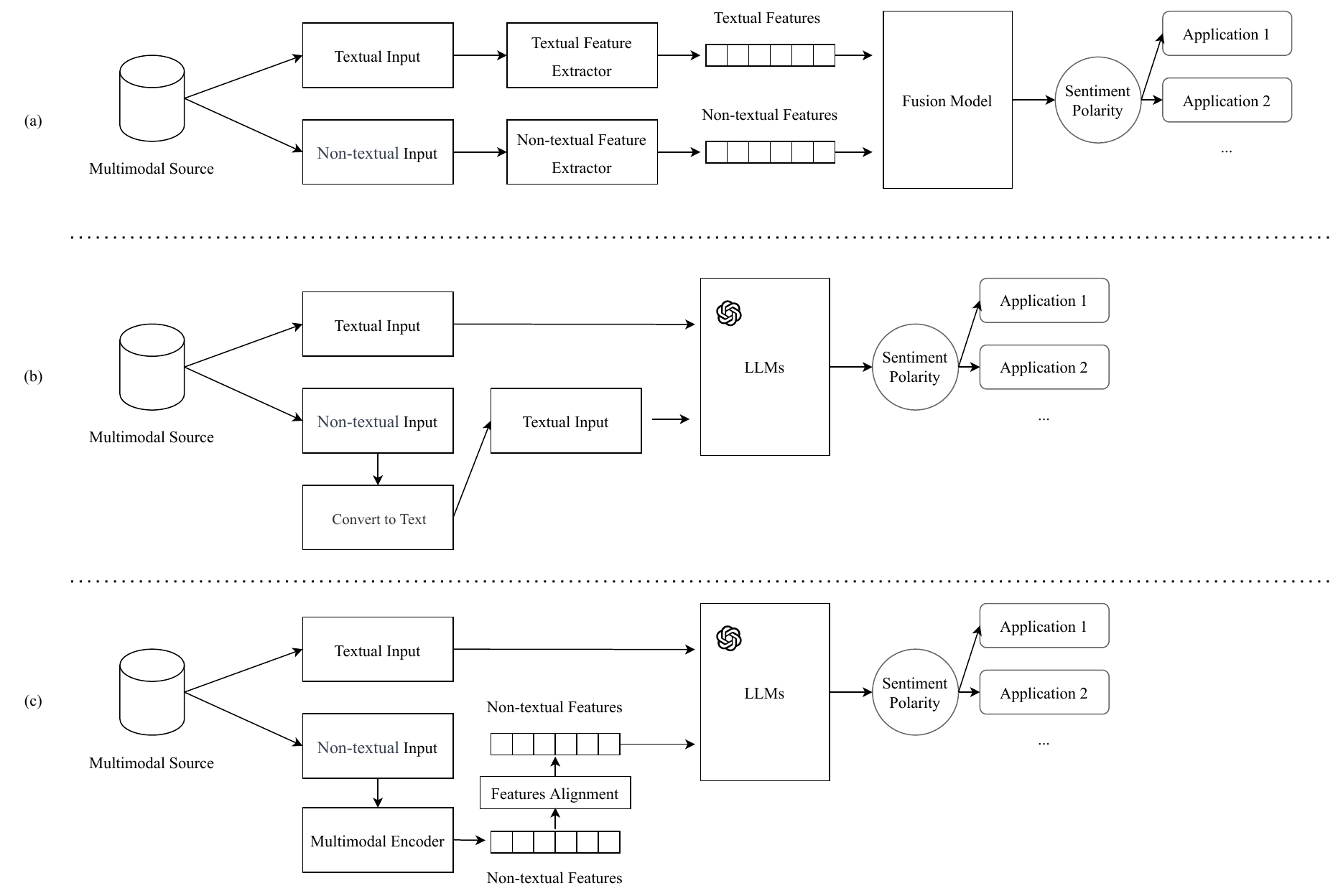}
    \caption{Conceptual illustration of multimodal sentiment analysis using LLMs: (a) multimodal sentiment analysis method without using LLMs, (b) modal transformation-based method, and (c) multimodal encoder-based method.}
    \label{fig:method} 
\end{figure*}

\subsection{The Usage of LLMs in Multimodal Sentiment Analysis}

Figure~\ref{fig:method} contrasts the workflow of multimodal sentiment analysis that leverages LLMs with the previous workflow. In the previous workflow, as shown in Figure~\ref{fig:method} (a), multimodal data was obtained from multimodal source like websites, and then being cleaned and selected. Various feature extraction algorithms are then employed to generate feature vectors for different modalities. Based on various multimodal fusion models and classification algorithms, sentiment prediction results were obtained and applied to specific applications. However, under LLMs, to leverage the rich knowledge and robust reasoning capabilities of LLMs and enable text-oriented LLMs to understand multimodal signals, there are primarily two strategies. One, as shown in Figure~\ref{fig:method} (b), is to textualize non-text inputs, allowing the LLMs to comprehend the textualized multimodal signals, such as converting images into text by image caption model. The other strategy, as shown in Figure~\ref{fig:method} (c), involves using a multimodal encoder to obtain multimodal features and then learning feature alignment mappings, aligning multimodal signals with text in the feature space, enabling LLMs to utilize multimodal features for sentiment analysis.

In Table~\ref{methods}  we have summarized some representative multimodal sentiment analysis methods assisted by LLMs, analyzing the strategies used with LLMs as well as their advantages and disadvantages. After analysis, we have found that most existing research tends to view LLMs as knowledge sources. Operating under a parameter-fixd paradigm, these studies leverage zero-shot and few-shot strategies to endow smaller models with additional worldly knowledge in multimodal sentiment analysis tasks, resulting in performance improvements. Here are further advantages and methods of utilizing LLMs in text-centric multimodal sentiment analysis:

\begin{itemize}
    \item LLMs can supplement richer knowledge, such as knowledge of different languages and cultures, to promote the progress of multimodal sentiment analysis towards multilingualism.
    \item Leveraging the robust multimodal capabilities of LMMs, models like GPT-4V and LLava, known for their strong image captioning abilities, can transform image data into textual format, simplifying the challenge of modal alignment.
    \item Utilizing the powerful reasoning capabilities of LLMs, existing work has shown that effective In- Context Learning (ICL) can enhance the emotional reasoning capabilities of LLMs, significantly improving their ability to trace and guide emotional understanding.
    \item Fine-tuning with high-quality multimodal sentiment data using a parameter-tuning paradigm, such as the A2II model, has also been successful. Although it used the smaller-scale Flan-T5-base model, there is anticipation for methods that adopt parameter-efficient fine-tuning strategy in larger-scale LLMs.
    \item Additionally, the use of LLMs as tools in multimodal sentiment analysis holds a promising outlook.
\end{itemize}

However, there are also disadvantages of using LLMs in multimodal sentiment analysis, including:

\begin{itemize}
    \item LLMs have to face hallucination problems, and the inevitable generation of erroneous knowledge may lead to incorrect judgments. Enhancing the accuracy and completeness of sentiment judgment-related knowledge from LLMs while reducing the negative noise caused by biases and hallucinations remains a pressing challenge.
    \item The sensitivity of LLMs to prompts is significant, as different prompts can drastically influence the output. Choosing the appropriate prompt is challenging.
    \item Not all LLMs excel in emotional intelligence; as the training of LLMs and LMMs currently aims to develop a broad range of capabilities, emotional intelligence is just one of many focal points. Therefore, the emotional capabilities of most models may not be exceptional, and careful consideration is needed when selecting LLMs for assisting in multimodal sentiment analysis.
    \item Existing LMMs still lack support for additional modalities. While most LMMs focus on text and image modalities, and some have video processing capabilities, there is a lack of capacity to handle other modalities like physiological signals, limiting their use in multimodal sentiment analysis.
    \item Methods based on the parameter-tuning paradigm face significant costs, requiring several times the computational resources and time compared to traditional multimodal sentiment analysis models.

\end{itemize}

\begin{figure*}[!htb]
    \centering
    \includegraphics[width=16cm]{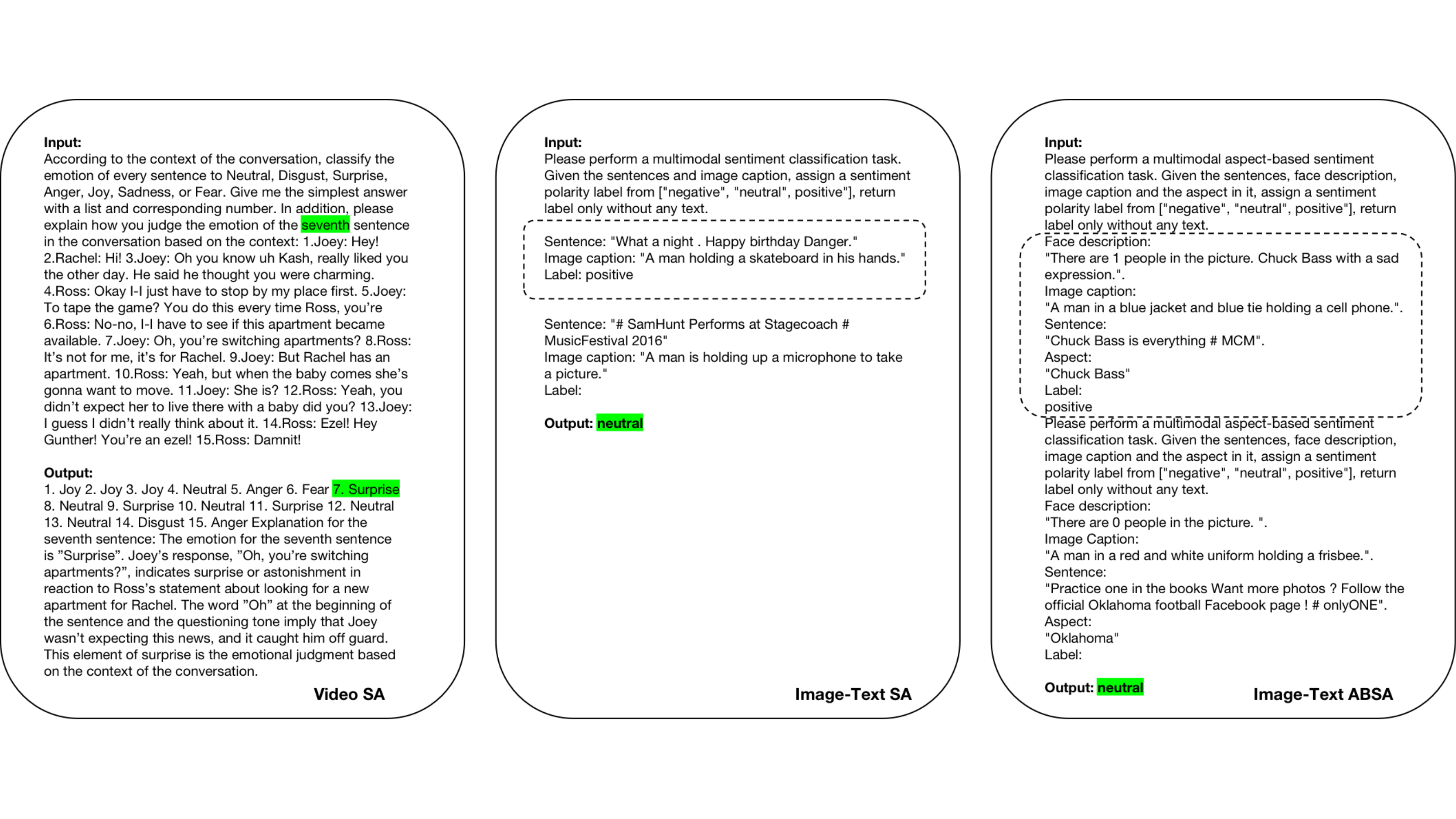}
    \caption{Prompt examples for video-based sentiment analysis (Video SA), image-text sentiment classification (Image-Text SA), and multimodal aspect-based sentiment classification(Image-Text ABSA) respectively. The text inside the dashed box are demonstration of the few-shot setting and would be removed under the zero-shot setting}
    \label{fig:prompt}
\end{figure*}

\section{Evaluations of LLMs-based Multimodal Sentiment Analysis Methods}\label{section4}


\subsection{Prompting Strategy}
When using LLMs, we employ prompts (a specific type of input text) to trigger the model's response. Since LLMs are highly sensitive to prompts, even slight variations in semantics can elicit vastly different responses. Therefore, prompt design is of paramount importance. Figure~\ref{fig:prompt} shows some prompt examples.

As shown in Figure~\ref{fig:prompt}, in the zero-shot setting, the prompts include the task name, task definition, and output format. The task name is used to identify and specify the task, while the task definition provides an explanation of the task, enabling the model to understand the input-output format of the task and providing a candidate label space for outputs. The output format defines the expected structure of the output, guiding the model to generate content in the expected format.

In the few-shot setting, additional demonstration sections are added to assist in model inference learning.

\subsection{Evaluation Metrics}
This section discusses the various commonly used metrics used in the
field of text-centric multimodal sentiment analysis tasks~\cite{215}.

\textbf{Accuracy} is a measure that indicates the proportion of instances correctly predicted out of the total number of instances. Further, \textbf{Weighted Accuracy} accounts for class imbalances by assigning different weights to each class.
\begin{gather}
    Accuracy = \frac{TP+TN}{TP+TN+FP+TN}, \\
    Weighted\text{-}Accuracy = \frac{1}{N} \sum_{i=1}^{N} w_{i}\cdot\frac{TP_{i}+TN_{i}}{TP_{i}+TN_{i}+FP_{i}+TN_{i}}
\end{gather}
where $TP$ represents True Positives, $TN$ represents True Negatives, $FP$ represents False Positives, $FN$ represents False Negatives, $N$ represents Total Number of Instances, $w_{i}$ represents Weight for Class $i$.

\textbf{Precision} evaluates the fraction of true positive predictions among instances that have been predicted as positive.
\begin{gather}
    Precision = \frac{TP}{TP+FP}
\end{gather}

\textbf{Recall}, which is sometimes referred to as Sensitivity or True Positive Rate, quantifies the proportion of true positive instances that are accurately predicted.
\begin{gather}
    Recall = \frac{TP}{TP+FN} 
\end{gather}

The \textbf{F1-Score} merges precision and recall to offer a well-rounded assessment of the model’s accuracy. Additionally, the \textbf{Weighted F1-Score} accounts for class imbalances.
\begin{gather}
    F1\text{-}Score = 2\cdot\frac{Precision\cdot Recall}{Precision+Recall}, \\
    Weighted\text{-}F1\text{-}Score = \frac{1}{N} \sum_{i=1}^{N} w_{i}\cdot 2\cdot\frac{Precision_{i}\cdot Recall_{i}}{Precision_{i}+Recall_{i}}
\end{gather}

\subsection{Reference Results}
With the in-depth development of LLMs in the field of multimodal sentiment analysis, it is necessary to compare the performance of LLMs on multimodal sentiment analysis datasets. However, testing on commercial LLMs such as ChatGPT is often expensive. Some works~\cite{141,160,171,206,207,213,214,217} have demonstrated the performance of some LLMs on multimodal sentiment analysis tasks, We have organized the relevant results in the table~\ref{reference_results}.

\begin{table}[h]
\caption{An overview of the performance of existing LLMs and LMMs on text-centric multimodal sentiment analysis benchmarks. $*$ indicates the model results after training on the MMSD and MMSD2.0 dataset.  \textit{Italicized words} represent the few-shot results, and the rest are zero-shot results. \textbf{Bold} represents the best zero-shot results.}
\scalebox{0.46}{
\begin{tabular}{l|ccc|cccc|cccc|cccccccc}
\hline
\textbf{Method}         & \textbf{MVSA-S} & \textbf{MVSA-M} & \textbf{TumEmo} & \multicolumn{2}{c}{\textbf{Twitter-2015}} & \multicolumn{2}{c|}{\textbf{Twitter-2017}} & \multicolumn{2}{c}{\textbf{MMSD}} & \multicolumn{2}{c|}{\textbf{MMSD2.0}} & \multicolumn{2}{c}{\textbf{MOSI-2}} & \multicolumn{2}{c}{\textbf{MOSEI}} & \multicolumn{2}{c}{\textbf{CH-SIMS}} & \multicolumn{2}{c}{\textbf{M3ED}} \\ \cline{2-20} 
\textbf{}               & \textbf{Acc.}   & \textbf{Acc.}   & \textbf{Acc.}   & \textbf{Acc.}         & \textbf{F1}       & \textbf{Acc.}         & \textbf{F1}        & \textbf{Acc.}     & \textbf{F1}   & \textbf{Acc.}      & \textbf{F1}      & \textbf{Acc.}      & \textbf{F1}    & \textbf{Acc.}     & \textbf{F1}    & \textbf{Acc.}      & \textbf{F1}     & \textbf{Acc.}    & \textbf{F1}    \\ \hline
\textbf{ChatGPT}        & 56.55           & 53.18           & 48.17           & 65.28                 & \textbf{63.12}             & 62.47                 & \textbf{62.34}              & 69.02             & -             & -                  & -                & 86.13              & 85.92          & 85.60             & 84.43          & 79.66              & 78.78           & 44.47            & \textbf{40.40}          \\
\textbf{Claude}         & -               & -               & -               & -                     & -                 & -                     & -                  & -                 & -             & -                  & -                & 87.04              & \textbf{86.55}          & 85.83             & \textbf{84.81}          & \textbf{88.70}              & \textbf{87.44}           & 34.90            & 34.83          \\
\textbf{LLaMA1-7B}      & 67.23           & 60.72           & 38.26           & 58.53                 & -                 & 46.43                 & -                  & 58.99             & -             & -                  & -                & 82.01              & -              & 75.62             & -              & -                  & -               & -                & -              \\
\textbf{LLaMA1-13B}     & 66.88           & 68.82           & 44.68           & 52.07                 & -                 & 47.24                 & -                  & 57.53             & -             & -                  & -                & 72.10              & -              & 79.55             & -              & -                  & -               & -                & -              \\
\textbf{LLaMA2-7B}      & 66.99           & 69.22           & 40.28           & 58.53                 & -                 & 46.60                 & -                  & 56.33             & -             & -                  & -                & 67.68              & -              & 77.30             & -              & -                  & -               & -                & -              \\
\textbf{LLaMA2-13B}     & 66.02           & 68.69           & 45.78           & 60.37                 & -                 & 48.54                 & -                  & 60.23             & -             & -                  & -                & 81.86              & -              & 81.66             & -              & -                  & -               & -                & -              \\
\textbf{Mixtral-AWQ}    & 54.37           & 55.59           & 43.94           & 55.45                 & -                 & 60.21                 & -                  & 64.38             & -             & -                  & -                & 87.92              & -              & 79.30             & -              & -                  & -               & -                & -              \\
\textbf{Gemma}          & 67.72           & 61.61           & 43.10           & 54.29                 & -                 & 52.43                 & -                  & 60.07             & -             & -                  & -                & 81.65              & -              & 77.05             & -              & -                  & -               & -                & -              \\
\textbf{Flan-T5-XXL}    & 64.81           & 66.01           & 49.56           & \textbf{72.13}        & -                 & 63.70                 & -                  & 71.40             & -             & -                  & -                & 89.60              & -              & 86.52             & -              & -                  & -               & -                & -              \\
\textbf{ChatGLM2-6B}    & -               & -               & -               & -                     & -                 & -                     & -                  & -                 & -             & -                  & -                & 84.12               & 84.12           & -                 & -              & 77.58               & 75.95            & \textbf{45.68}             & 30.52           \\
\textbf{ChatGLM2-6B*}   & -               & -               & -               & -                     & -                 & -                     & -                  & 94.02             & 93.76         & 78.41              & 78.23            & -                  & -              & -                 & -              & -                  & -               & -                & -              \\
\textbf{LLaMA2-7B*}     & -               & -               & -               & -                     & -                 & -                     & -                  & 93.97             & 93.72         & 82.52              & 82.27            & -                  & -              & -                 & -              & -                  & -               & -                & -              \\ \hline
\textbf{GPT-4V}         & 76.19           & \textbf{71.05}  & 50.58           & 53.85                 & -                 & 60.16                 & -                  & \textbf{76.76}    & -             & -                  & -                & \textbf{90.91}     & -              & 87.10             & -              & -                  & -               & -                & -              \\
\textbf{Claude3-V}      & \textbf{80.95}  & 69.08           & 46.15           & 38.46                 & -                 & 54.47                 & -                  & 71.37             & -             & -                  & -                & 78.79              & -              & 79.93             & -              & -                  & -               & -                & -              \\
\textbf{Gemini-V}       & 72.73           & 70.18           & 51.65           & 54.51                 & -                 & 59.32                 & -                  & 56.83             & -             & -                  & -                & 88.34              & -              & \textbf{87.14}    & -              & -                  & -               & -                & -              \\
\textbf{OpenFlamingo}   & 55.58           & 61.15           & 29.47           & 57.28                 & -                 & 46.19                 & -                  & 52.68             & -             & -                  & -                & 79.97              & -              & 77.30             & -              & -                  & -               & -                & -              \\
\textbf{Fromage}        & 29.85           & 28.19           & 22.76           & 19.96                 & -                 & 27.31                 & -                  & 40.68             & -             & -                  & -                & 57.19              & -              & 47.41             & -              & -                  & -               & -                & -              \\
\textbf{LLaVA-v0-7B}    & 69.42           & 65.42           & 30.44           & 35.10                 & -                 & 44.57                 & -                  & 43.21             & -             & -                  & -                & 74.69              & -              & 74.65             & -              & -                  & -               & -                & -              \\
\textbf{LLaVA-v0-13B}   & 73.06           & 69.61           & 38.51           & 37.99                 & -                 & 48.46                 & -                  & 44.29             & -             & -                  & -                & 80.18              & -              & 76.58             & -              & -                  & -               & -                & -              \\
\textbf{LLaVA-v1.6-7B}  & 59.95           & 67.23           & 45.12           & 59.31                 & -                 & 52.84                 & -                  & 59.61             & -             & -                  & -                & 85.63              & -              & 81.62             & -              & -                  & -               & -                & -              \\
\textbf{LLaVA-v1.6-13B} & 64.56           & 60.43           & 53.22           & 58.73                 & -                 & 56.08                 & -                  & 62.31             & -             & -                  & -                & 86.39              & -              & 78.26             & -              & -                  & -               & -                & -              \\
\textbf{MiniGPT4}       & 71.12           & 70.78           & 50.29           & 47.16                 & -                 & 49.43                 & -                  & 57.49             & -             & -                  & -                & 83.99              & -              & 83.38             & -              & -                  & -               & -                & -              \\
\textbf{mPLOG-Owl}      & 51.94           & 50.36           & 33.37           & 33.75                 & -                 & 38.74                 & -                  & 49.73             & -             & -                  & -                & 68.75              & -              & 58.10             & -              & -                  & -               & -                & -              \\
\textbf{mPLOG-Owl2.1}   & 53.64           & 63.11           & 47.02           & 60.66                 & -                 & 55.11                 & -                  & 60.48             & -             & -                  & -                & 85.63              & -              & 73.47             & -              & -                  & -               & -                & -              \\
\textbf{AdapterV2}      & 73.54           & 70.13           & 39.14           & 37.32                 & -                 & 48.38                 & -                  & 57.20             & -             & -                  & -                & 86.43              & -              & 82.02             & -              & -                  & -               & -                & -              \\
\textbf{VPGTrans}       & 64.32           & 69.54           & 46.17           & 42.62                 & -                 & 44.81                 & -                  & 65.04             & -             & -                  & -                & 76.22              & -              & 76.76             & -              & -                  & -               & -                & -              \\
\textbf{MultiGPT}       & 52.91           & 62.03           & 30.26           & 58.53                 & -                 & 46.35                 & -                  & 59.82             & -             & -                  & -                & 68.35              & -              & 72.76             & -              & -                  & -               & -                & -              \\
\textbf{LaVIN-7B}       & 39.32           & 40.75           & 26.84           & 37.22                 & -                 & 33.06                 & -                  & 60.48             & -             & -                  & -                & 71.41              & -              & 69.97             & -              & -                  & -               & -                & -              \\
\textbf{LaVIN-13B}      & 53.64           & 48.79           & 32.77           & 35.39                 & -                 & 40.68                 & -                  & 57.58             & -             & -                  & -                & 79.97              & -              & 73.54             & -              & -                  & -               & -                & -              \\
\textbf{Lynx}           & 64.32           & 67.71           & 42.79           & 46.00                 & -                 & 47.00                 & -                  & 43.96             & -             & -                  & -                & 74.77              & -              & 73.72             & -              & -                  & -               & -                & -              \\
\textbf{Fuyu-8B}        & 48.54           & 55.46           & 46.34           & 58.82                 & -                 & 50.81                 & -                  & 61.44             & -             & -                  & -                & 83.49              & -              & 78.37             & -              & -                  & -               & -                & -              \\
\textbf{LaVIT}          & 61.65           & 68.74           & 41.78           & 36.84                 & -                 & 43.36                 & -                  & 56.00             & -             & -                  & -                & 73.09              & -              & 64.10             & -              & -                  & -               & -                & -              \\
\textbf{Qwen-VL-Chat}   & 62.38           & 69.06           & 49.29           & 65.48                 & -                 & 59.72                 & -                  & 61.10             & -             & -                  & -                & 85.93              & -              & 80.41             & -              & -                  & -               & -                & -              \\
\textbf{BLIP}           & 66.26           & 68.22           & 51.06           & 70.78                 & -                 & \textbf{64.42}        & -                  & 72.02             & -             & -                  & -                & 88.99              & -              & 86.88             & -              & -                  & -               & -                & -              \\
\textbf{InstructBLIP}   & 71.60           & 70.37           & \textbf{52.36}  & 57.57                 & 59.63             & 60.37                 & 35.96              & 73.10             & -             & -                  & -                & 88.68              & -              & 85.98             & -              & -                  & -               & -                & -              \\
\textbf{LLaVA-v1.5-7B*} & -               & -               & -               & -                     & -                 & -                     & -                  & 93.67             & 93.40         & 85.18              & 85.11            & -                  & -              & -                 & -              & -                  & -               & -                & -              \\
\textbf{mPLUG-Owl2}     & -               & -               & -               & \textit{76.80}                 & \textit{72.30}             & \textit{74.20}                 & \textit{73.00}              & -                 & -             & -                  & -                & -                  & -              & -                 & -              & -                  & -               & -                & -              \\
\textbf{MMICL-14B}      & -               & -               & -               & \textit{76.00}                 & \textit{72.70}             & \textit{74.10}                 & \textit{74.00}              & -                 & -             & -                  & -                & -                  & -              & -                 & -              & -                  & -               & -                & -              \\
\textbf{LLaVA-v1.5-13B} & -               & -               & -               & \textit{77.90}                 & \textit{74.30}             & \textit{74.60}                 & \textit{74.30}              & -                 & -             & 51.06              & 43.02            & -                  & -              & -                 & -              & -                  & -               & -                & -              \\
\textbf{Qwen-VL-v1.0}   & -               & -               & -               & -                     & -                 & -                     & -                  & -                 & -             & \textbf{76.63}              & \textbf{69.03}            & -                  & -              & -                 & -              & -                  & -               & -                & -              \\ \hline
\end{tabular}
}
\label{reference_results}

\end{table}

\section{Applications of text-centric multimodal sentiment analysis}\label{section5}
The research in text-centric multimodal sentiment analysis has its roots in the flourishing development of multimodal data and the advancements in deep learning technologies. It is also driven by a wide range of practical applications. In this section we explore the application of LLM-based text-centric multimodal sentiment analysis.

\subsection{Comment analysis}
One of the earliest and most impactful applications of sentiment analysis was in the field of e-commerce for comment analysis. This research area not only attracted numerous computer scientists who delved into algorithm development but also drew the interest of management scientists exploring marketing and management strategies. Initially, these studies primarily revolved around textual comments, analyzing user reviews to gather feedback on products or services. However, as e-commerce evolved, relying solely on text-based sentiment analysis proved insufficient. User-generated comments often include multimedia elements, making multimodal data more prominent compared to pure text comments.

With the increasing availability of multimodal data on social networks, some of the challenges that puzzled researchers can be alleviated in a multimodal interactive context, enabling comprehensive sentiment analysis. For instance, one challenging problem is sarcasm recognition, which can be easily resolved with the addition of multimodal information. For example, when a comment like ``It's such a surprise'' is accompanied by a picture of a disappointed face, sarcasm recognition becomes straightforward. In the field of management, multimodal data, enriched with additional modal factors, can influence user decisions and consequently impact marketing and management strategies.

In practical applications, fine-grained sentiment analysis is more effective. In text-based analysis studies, user textual comment data can be broken down into fine-grained segments (e.g., sentences, clauses), with each segment evaluating different aspects of the main entity (e.g., price, quality, appearance). In contrast, fine-grained analysis of multimodal data is still in its emerging stages but presents greater challenges. For instance, extracting object-level information from complex multimodal data and modeling fine-grained element correspondences between multimodal elements are ongoing research topics that need exploration with LLMs in the future.

\subsection{Multimodal intelligent human-machine interaction}
Multimodal sentiment analysis can be applied to human-computer interaction, enabling real-time understanding and analysis of emotional communication for more natural interactions. There are three main categories of applications: 

\textbf{1) Customer Service Conversations.} In this domain, multimodal data consists of audio data and text data transformed from Automatic Speech Recognition (ASR) technology. It mainly serves two tasks: customer satisfaction analysis and detection of customer abnormal emotions. Customer satisfaction analysis involves using multimodal emotion computing technology to analyze the content of conversations between customers and service representatives to assess the level of customer satisfaction. Customer abnormal emotion detection monitors customer emotions in real time through the analysis of customer dialogue data and prompts timely intervention when abnormal emotional changes occur.

\textbf{2) Emotional Companionship.} Emotional companionship is a crucial aspect of chatbot applications. Currently, most companion chatbots do not utilize multimodal emotion computing technology, meaning they do not fully possess human-like multimodal processing capabilities. Ideally, companion chatbots should be capable of recognizing and generating multimodal emotional features, such as expressing emotions through language, exhibiting emotional fluctuations in speech, or displaying facial expressions.

\textbf{3) Smart Furniture}
The development of artificial intelligence has given rise to smart homes, which enhance convenience and comfort in daily life and are increasingly popular among consumers. Many tech companies worldwide have entered the smart home market, proposing a range of solutions like Apple's HomeKit, Xiaomi's Mi Home, and Haier's U-Home. While smart homes have made life more convenient, they are currently primarily focused on home automation, with users controlling home devices through voice commands based on keyword recognition. This approach does not fully embody the intelligence of smart homes, and there is substantial potential for further development, particularly in voice interaction and automatic environment detection.

With the assistance of LLMs, AI technologies based on multimodal sentiment analysis methods can elevate the intelligence of smart homes in the future. True smart home scenarios involve multiple modalities, where smart home products can provide appropriate feedback by calculating the user's emotions (e.g., happiness, anger, sadness) or states (e.g., fatigue, restlessness). For example, based on a user's fatigue state in a multimodal scenario, the system could ask if they want the lights dimmed. In conversational scenarios, the system can detect the user's emotional state and provide empathetic responses. Smart in-car systems can promptly detect abnormal user emotions or states (e.g., road rage, fatigue) and provide appropriate reminders. Designing these functionalities poses significant challenges, but it also represents a significant opportunity for multimodal emotion computing to enter the smart furniture domain.

\section{Conclusions}\label{section6}
In this survey, we introduced the latest advancements in text-centric multimodal sentiment analysis area and summarized the primary challenges and potential solutions. Additionally, we reviewed the existing ways of applying LLMs in multimodal sentiment analysis tasks and summarized their advantages and disadvantages. We believe that leveraging LLMs in multimodal sentiment analysis has several potential advantages: 1) Knowledge Source: LLMs trained on massive datasets and can be treated as a knowledge source, that can capture a broader range of patterns, linguistic cues, and contextual information related to emotions, potentially improving recognition performance. 2) Interpretability: LLMs can potentially elucidate the reasoning behind their decisions, enhancing the interpretability and transparency of the emotion recognition process. 3)  Cross-Domain Applications: LLMs have the potential to be applied across various domains, as they are trained on a wide range of data sources. This allows them to understand emotions expressed in various domains, from customer reviews to conversational data, thus achieving broader applicability. However, LLMs-based methods also have to face problems such as hallucinations and high fine-tuning costs. The emergence of LLMs provides new ideas and challenges for multimodal sentiment analysis. We hope that this survey can help and encourage further research in this field.

\newpage


\begin{thebibliography}{99}
\bibitem{1}Hatzivassiloglou V, McKeown KR. Predicting the semantic orientation of adjectives. In{\it Proc. of the EACL'97. Morristown: ACL}, 1997. 174-181.
\bibitem{2} Tom B. Brown, Benjamin Mann, Nick Ryder, Melanie Subbiah, Jared Kaplan, Prafulla Dhariwal, Arvind Neelakantan, Pranav Shyam, Girish Sastry, Amanda Askell, Sandhini Agarwal, Ariel Herbert-Voss, Gretchen Krueger, Tom Henighan, Rewon Child, Aditya Ramesh, Daniel M. Ziegler, Jeffrey Wu, Clemens Winter, Christopher Hesse, Mark Chen, Eric Sigler, Mateusz Litwin, Scott Gray, Benjamin Chess, Jack Clark, Christopher Berner, Sam McCandlish, Alec Radford, Ilya Sutskever, and Dario Amodei. 2020. Language models are few-shot learners. In {\it Advances in Neural Information Processing Systems 33: Annual Conference on Neural Information Processing Systems 2020, NeurIPS 2020}, December 6-12, 2020, virtual.

\bibitem{3} Jack W. Rae, Sebastian Borgeaud, Trevor Cai, Katie Millican, Jordan Hoffmann, H. Francis Song, John Aslanides, Sarah Henderson, Roman Ring, Susannah Young, Eliza Rutherford, Tom Hennigan, Jacob Menick, Albin Cassirer, Richard Powell, George van den Driessche, Lisa Anne Hendricks, Maribeth Rauh, Po-Sen Huang, Amelia Glaese, Johannes Welbl, Sumanth Dathathri, Saffron Huang, Jonathan Uesato, John Mellor, Irina Higgins, Antonia Creswell, Nat McAleese, Amy Wu, Erich Elsen, Siddhant M. Jayakumar, Elena Buchatskaya, David Budden, Esme Sutherland, Karen Simonyan, Michela Paganini, Laurent Sifre, Lena Martens, Xiang Lorraine Li, Adhiguna Kuncoro, Aida Nematzadeh, Elena Gribovskaya, Domenic Donato, Angeliki Lazaridou, Arthur Mensch, JeanBaptiste Lespiau, Maria Tsimpoukelli, Nikolai Grigorev, Doug Fritz, Thibault Sottiaux, Mantas Pajarskas, Toby Pohlen, Zhitao Gong, Daniel Toyama, Cyprien de Masson d'Autume, Yujia Li, Tayfun Terzi, Vladimir Mikulik, Igor Babuschkin, Aidan Clark, Diego de Las Casas, Aurelia Guy, Chris Jones, James Bradbury, Matthew J. Johnson Blake A. Hechtman, LauraWeidinger, Iason Gabriel, William S. Isaac, Edward Lockhart, Simon Osindero, Laura Rimell, Chris Dyer, Oriol Vinyals, Kareem Ayoub, Jeff Stanway, Lorrayne Bennett, Demis Hassabis, Koray Kavukcuoglu, and Geoffrey Irving. 2021. Scaling language models: Methods, analysis \& insights from training gopher. {\it CoRR}, abs/2112.11446.

\bibitem{4} OpenAI. 2023. GPT-4 technical report. {\it CoRR} , abs/2303.08774.

\bibitem{5} Chao Zhang, Zichao Yang, Xiaodong He and Li Deng. Multimodal Intelligence: Representation Learning, Information Fusion, and Applications. {\it IEEE JOURNAL OF SELECTED TOPICS IN SIGNAL PROCESSING}. 2020, 14(3): 478-493.

\bibitem{6} Xiang Deng, Vasilisa Bashlovkina, Feng Han, Simon Baumgartner, and Michael Bendersky. 2023. Llms to the moon? reddit market sentiment analysis with LLMs. In {\it Companion Proceedings of the ACM Web Conference 2023, WWW2023}, pages 1014-1019.

\bibitem{7} Qihuang Zhong, Liang Ding, Juhua Liu, Bo Du, and Dacheng Tao. 2023. Can chatgpt understand too? A comparative study on chatgpt and fine-tuned BERT. {\it CoRR}, abs/2302.10198.

\bibitem{8} Zengzhi Wang, Qiming Xie, Zixiang Ding, Yi Feng, and Rui Xia. 2023. Is chatgpt a good sentiment analyzer? A preliminary study. {\it CoRR}, abs/2304.04339.

\bibitem{9} Zhang W, Deng Y, Liu B, et al. Sentiment Analysis in the Era of Large Language Models: A Reality Check[J]. arXiv preprint arXiv:2305.15005, 2023.

\bibitem{10} Yejin Bang, Samuel Cahyawijaya, Nayeon Lee, Wenliang Dai, Dan Su, Bryan Wilie, Holy Lovenia, Ziwei Ji, Tiezheng Yu, Willy Chung, Quyet V. Do, Yan Xu, and Pascale Fung. 2023. A multitask, multilingual, multimodal evaluation of chatgpt on reasoning, hallucination, and interactivity. {\it CoRR}, abs/2302.04023.

\bibitem{11} Junjie Ye, Xuanting Chen, Nuo Xu, Can Zu, Zekai Shao, Shichun Liu, Yuhan Cui, Zeyang Zhou, Chao Gong, Yang Shen, Jie Zhou, Siming Chen, Tao Gui, Qi Zhang, and Xuanjing Huang. 2023. A comprehensive capability analysis of GPT-3 and GPT-3.5 series models. {\it CoRR}, abs/2303.10420

\bibitem{12} Jingfeng Yang, Hongye Jin, Ruixiang Tang, Xiaotian Han, Qizhang Feng, Haoming Jiang, Bing Yin, and Xia Hu. 2023. Harnessing the power of llms in practice: A survey on chatgpt and beyond. {\it CoRR}, abs/2304.13712.

\bibitem{13} Richard Socher, Alex Perelygin, Jean Wu, Jason Chuang, Christopher D. Manning, Andrew Y. Ng, and Christopher Potts. 2013. Recursive deep models for semantic compositionality over a sentiment treebank. In {\it Proceedings of the 2013 Conference on Empirical Methods in Natural Language Processing, EMNLP 2013, 18-21 October 2013, Grand Hyatt Seattle, Seattle, Washington, USA, A meeting of SIGDAT, a Special Interest Group of the ACL}, pages 1631-1642.

\bibitem{14} Tadas Baltrus aitis, Chaitanya Ahuja, and Louis-Philippe Morency. Multimodal machine learning: A survey and taxonomy. {\it IEEE transactions on pattern analysis and machine intelligence.} 2018, 41(2): 423-443.

\bibitem{15} Paul Mc Kevitt. MultiModal Semantic Representation[C]// Tilburg University.  {\it First Working Meeting of the SIGSEM Working Group on the Representation of MultiModal Semantic Information.} Tilburg: Tilburg University, 2003: 1-16.

\bibitem{16} Yi Tay, Mostafa Dehghani, Vinh Q. Tran, Xavier Garcia, Dara Bahri, Tal Schuster, Huaixiu Steven Zheng, Neil Houlsby, and Donald Metzler. 2022. Unifying language learning paradigms. {\it CoRR}, abs/2205.05131.

\bibitem{17} Hugo Touvron, Thibaut Lavril, Gautier Izacard, Xavier Martinet, Marie-Anne Lachaux, Timothee Lacroix, Baptiste Roziere, Naman Goyal, Eric Hambro, Faisal
Azhar, Aurelien Rodriguez, Armand Joulin, Edouard Grave, and Guillaume Lample. 2023. Llama: Open and efficient foundation language models. {\it CoRR}, abs/2302.13971.

\bibitem{18} Zhu D, Chen J, Shen X, et al. Minigpt-4: Enhancing vision-language understanding with advanced large language models[J]. arXiv preprint arXiv:2304.10592, 2023.

\bibitem{19} Liu H, Li C, Wu Q, et al. Visual instruction tuning[J]. arXiv preprint arXiv:2304.08485, 2023.

\bibitem{20} Wenliang Dai and Junnan Li and Dongxu Li and Anthony Meng Huat Tiong and Junqi Zhao and Weisheng Wang and Boyang Li and Pascale Fung and Steven Hoi. InstructBLIP: Towards General-purpose Vision-Language Models with Instruction Tuning[J]. arXiv preprint arXiv:2305.06500, 2023.

\bibitem{21} https://www.bing.com.

\bibitem{22} Aakanksha Chowdhery, Sharan Narang, Jacob Devlin, Maarten Bosma, Gaurav Mishra, Adam Roberts, Paul Barham, Hyung Won Chung, Charles Sutton, Sebastian Gehrmann, Parker Schuh, Kensen Shi, Sasha Tsvyashchenko, Joshua Maynez, Abhishek Rao, Parker Barnes, Yi Tay, Noam Shazeer, Vinodkumar Prabhakaran, Emily Reif, Nan Du, Ben Hutchinson, Reiner Pope, James Bradbury, Jacob Austin, Michael Isard, Guy Gur-Ari, Pengcheng Yin, Toju Duke, Anselm Levskaya, Sanjay Ghemawat, Sunipa Dev, Henryk Michalewski, Xavier Garcia, Vedant Misra, Kevin Robinson, Liam Fedus, Denny Zhou, Daphne Ippolito, David Luan, Hyeontaek Lim, Barret Zoph, Alexander Spiridonov, Ryan Sepassi, David Dohan, Shivani Agrawal, Mark Omernick, Andrew M. Dai, Thanumalayan Sankaranarayana Pillai, Marie Pellat, Aitor Lewkowycz, Erica Moreira, Rewon Child, Oleksandr Polozov, Katherine Lee, Zongwei Zhou, Xuezhi Wang, Brennan Saeta, Mark Diaz, Orhan Firat, Michele Catasta, Jason Wei, Kathy Meier-Hellstern, Douglas Eck, Jeff Dean, Slav Petrov, and Noah Fiedel. 2022. Palm: Scaling language modeling with pathways. {\it CoRR}, abs/2204.02311.

\bibitem{23} Taylor R, Kardas M, Cucurull G, et al. Galactica: A large language model for science[J]. arXiv preprint arXiv:2211.09085, 2022.

\bibitem{24} Meta A I. Introducing LLaMA: A foundational, 65-billion-parameter large language model[J]. Meta AI. https://ai. facebook. com/blog/large-language-model-llama-meta-ai, 2023.

\bibitem{25} Jason Wei, Maarten Bosma, Vincent Y. Zhao, Kelvin Guu, Adams Wei Yu, Brian Lester, Nan Du, Andrew M. Dai, and Quoc V. Le. 2022a. Finetuned language models are zero-shot learners. In {\it The Tenth International Conference on Learning Representations, ICLR 2022}, Virtual Event, April 25-29, 2022. OpenReview.net.

\bibitem{26} Victor Sanh, Albert Webson, Colin Raffel, Stephen H. Bach, Lintang Sutawika, Zaid Alyafeai, Antoine Chaffin, Arnaud Stiegler, Arun Raja, Manan Dey, M Saiful Bari, Canwen Xu, Urmish Thakker, Shanya Sharma Sharma, Eliza Szczechla, Taewoon Kim, Gunjan Chhablani, Nihal V. Nayak, Debajyoti Datta, Jonathan Chang, Mike Tian-Jian Jiang, Han Wang, Matteo Manica, Sheng Shen, Zheng Xin Yong, Harshit Pandey, Rachel Bawden, Thomas Wang, Trishala Neeraj, Jos Rozen, Abheesht Sharma, Andrea Santilli, Thibault Fevry, Jason Alan Fries, Ryan Teehan, Teven Le Scao, Stella Biderman, Leo Gao, Thomas Wolf, and Alexander M. Rush. 2022. Multitask-prompted training enables zero-shot task generalization. In {\it The Tenth International Conference on Learning Representations, ICLR 2022}, Virtual Event, April 25-29, 2022. OpenReview.net.

\bibitem{27} Hyung Won Chung, Le Hou, Shayne Longpre, Barret Zoph, Yi Tay, William Fedus, Eric Li, Xuezhi Wang, Mostafa Dehghani, Siddhartha Brahma, Albert Webson, Shixiang Shane Gu, Zhuyun Dai, Mirac Suzgun, Xinyun Chen, Aakanksha Chowdhery, Sharan Narang, Gaurav Mishra, Adams Yu, Vincent Y. Zhao, Yanping Huang, Andrew M. Dai, Hongkun Yu, Slav Petrov, Ed H. Chi, Jeff Dean, Jacob Devlin, Adam Roberts, Denny Zhou, Quoc V. Le, and Jason Wei. 2022. Scaling instruction-finetuned language models. {\it CoRR}, abs/2210.11416.

\bibitem{28} Zhang Z, Peng L, Pang T, et al. Refashioning Emotion Recognition Modelling: The Advent of Generalised Large Models[J]. arXiv preprint arXiv:2308.11578, 2023.

\bibitem{29} Cao D, Ji R, Lin D, et al. A Cross-media Public Sentiment Analysis System for Microblog[J]. {\it Multimedia Systems}, 2016, 22(4): 479-486.

\bibitem{30} You Q, Cao L, Jin H, et al. Robust Visual-textual Sentiment Analysis: When Attention Meets Tree-structured Recursive Neural Networks[C]// In {\it Proceedings of the 24th ACM international conference on multimedia}. New York: ACM, 2016: 1008-1017.

\bibitem{31} Truong Q T, Lauw H W. Vistanet: Visual Aspect Attention Network for Multimodal Sentiment Analysis[C]// In {\it Proceedings of the AAAI Conference on Artificial Intelligence. Palo Alto, California USA: AAAI Press}, 2019, 33(1): 305-312.
\bibitem{32} Khan Z, Fu Y. Exploiting BERT For Multimodal Target Sentiment Classification Through Input Space Translation[C]// In {\it Proceedings of the 29th ACM International Conference on Multimedia}. 2021: 3034-3042.

\bibitem{33} Zadeh A, Chen M, Poria S, et al. Tensor Fusion Network for Multimodal Sentiment Analysis[C] // In {\it Proceedings of the 2017 Conference on Empirical Methods in Natural Language Processing. [S.l.]: Association for Computational Linguistics}, 2017: 1103-1114. 

\bibitem{34} Wang Y, Shen Y, Liu Z, et al. Words can shift: Dynamically adjusting word representations using nonverbal behaviors[C]// In {\it Proceedings of the AAAI Conference on Artificial Intelligence}. 2019, 33(01): 7216-7223.

\bibitem{35} Hazarika D, Zimmermann R, Poria S. MISA: Modality-Invariant and -Specific Representations for Multimodal Sentiment Analysis // In {\it Proceedings of the 28th ACM International Conference on Multimedia. New York: ACM}, 2020: 1122-1131. 

\bibitem{36} Chen M, Wang S, Liang P P, et al. Multimodal Sentiment Analysis with WordLevel Fusion and Reinforcement Learning // In {\it Proceedings of the 19th ACM International Conference on Multimodal Interaction. New York: ACM}, 2017: 163-171.
\bibitem{37} Rahman W, Hasan M K, Lee S, et al. Integrating Multimodal Information in LargePretrained Transformers[C]// In {\it Proceedings of the 58th Annual Meeting of the Association for Computational Linguistics. [S.l.]: Association for Computational Linguistics}, 2020: 2359-2369. 

\bibitem{38} Li L, Chen Y C, Cheng Y, et al. Hero: Hierarchical Encoder for Video+ Language Omni-representation Pre-training[C]// In {\it Proceedings of the 2020 Conference on Empirical Methods in Natural Language Processing (EMNLP). [S.l.]: Association for Computational Linguistics}, 2020: 2046-2065.
\bibitem{39} Cao D, Ji R, Lin D, et al. A Cross-media Public Sentiment Analysis System for Microblog[J]. {\it Multimedia Systems}, 2016, 22(4): 479-486.

\bibitem{40} Yu W, Xu H, Yuan Z, et al. Learning modality-specific representations with self-supervised multi-task learning for multimodal sentiment analysis[J]. arXiv preprint arXiv:2102.04830, 2021.

\bibitem{41} Wu Y, Zhao Y, Yang H, et al. Sentiment Word Aware Multimodal Refinement for Multimodal Sentiment Analysis with ASR Errors[J]. arXiv preprint arXiv:2203.00257, 2022.

\bibitem{42} Liang B, Lou C, Li X, et al. Multi-Modal Sarcasm Detection with Interactive In-Modal and Cross-Modal Graphs[C]// In {\it Proceedings of the 29th ACM International Conference on Multimedia}. 2021: 4707-4715.

\bibitem{43} Tsai Y H H, Bai S, Liang P P, et al. Multimodal transformer for unaligned multimodal language sequences[C]// {\it Association for Computational Linguistics. Proceedings of the 57th Annual Meeting of the Association for Computational Linguistics. Florence, Italy: Association for Computational Linguistics}, 2019: 6558-6569.

\bibitem{44} Torres, E.P.; Torres, E.A.; Hernandez-Alvarez, M.; Yoo, S.G. EEG-Based BCI Emotion Recognition: A Survey. Sensors 2020, 20, 5083. https://doi.org/10.3390/s20185083.

\bibitem{45} Xiao-Wei Wang, Dan Nie, and Bao-Liang Lu. Eeg-based emotion recognition using frequency domain features and support vector machines. In International conference on neural information processing, pages 734-743. Springer, 2011.

\bibitem{46} Fatemeh Bahari and Amin Janghorbani. Eeg-based emotion recognition using recurrence plot analysis and k nearest neighbor classifier. In {\it2013 20th Iranian Conference on Biomedical Engineering (ICBME)}, pages 228-233. IEEE, 2013.
\bibitem{47} Jia, Ziyu \& Lin, Youfang \& Cai, Xiyang \& Chen, Haobin \& Gou, Haijun \& Wang, Jing. (2020). SST-EmotionNet: Spatial-Spectral-Temporal based Attention 3D Dense Network for EEG Emotion Recognition. 2909-2917. 10.1145/3394171.3413724.

\bibitem{48} Deng, Xiangwen \& Yang, Shangming \& Zhu, Junlin. (2021). SFE-Net: EEG-based Emotion Recognition with Symmetrical Spatial Feature Extraction.

\bibitem{49} Riloff E, Qadir A, Surve P, et al. Sarcasm as Contrast between a Positive Sentiment and Negative Situation[C]// {\it Proceedings of the 2013 Conference on Empirical Methods in Natural Language Processing. [S.l.]: Association for Computational Linguistics}, 2013: 704-714.
\bibitem{50} Yi Tay, Anh Tuan Luu, Siu Cheung Hui, and Jian Su. Reasoning with sarcasm by reading in between. Proceedings of the 56th Annual Meeting of the Association for Computational Linguistics. 2018: 1010-1020.
\bibitem{51} Xiong T, Zhang P, Zhu H, et al. Sarcasm Detection with Self-matching Networks and Low-Rank Bilinear Pooling[C]// The World Wide Web Conference. New York: ACM, 2019: 2115-2124.
\bibitem{52} Cheang H S, Pell M D. The Sound of Sarcasm[J]. Speech Communication, 2008, 50(5): 366-381.
\bibitem{53} Santiago Castro, Devamanyu Hazarika, Veronica PerezRosas, Roger Zimmermann, Rada Mihalcea, and Soujanya Poria. Towards multimodal sarcasm detection (an obviously perfect paper). Proceedings of the 57th Annual Meeting of the Association for Computational Linguistics. 2019: 4619-4629. 
\bibitem{54} Dushyant Singh Chauhan, SR Dhanush, Asif Ekbal, and Pushpak Bhattacharyya. Sentiment and emotion help sarcasm? a multi-task learning framework for multi-modal sarcasm, sentiment and emotion analysis. Proceedings of the 58th Annual Meeting of the Association for Computational Linguistics. 2020: 4351-4360.
\bibitem{55} Yitao Cai, Huiyu Cai, and Xiaojun Wan. Multimodal sarcasm detection in twitter with hierarchical fusion model. Proceedings of the 57th Annual Meeting of the Association for Computational Linguistics. 2019: 2506-2515.
\bibitem{56} Tang D, Qin B, Feng X, et al. Effective LSTMs for target-dependent sentiment classification[J]. arXiv preprint arXiv:1512.01100, 2015.
\bibitem{57} Wu, Yang \& Zhao, Yanyan \& Lu, Xin \& Qin, Bing \& Wu, Yin \& Sheng, Jian \& Li, Jinlong. (2021). Modeling Incongruity between Modalities for Multimodal Sarcasm Detection. IEEE MultiMedia. PP. 1-1. 10.1109/MMUL.2021.3069097.
\bibitem{58}Jennifer Woodland and Daniel Voyer. Context and intonation in the perception of sarcasm. Metaphor and Symbol. 2011, 26(3):227-239.
\bibitem{59} Ma D, Li S, Zhang X, et al. Interactive attention networks for aspect-level sentiment classification[J]. arXiv preprint arXiv:1709.00893, 2017.
\bibitem{60} Zhang C, Li Q, Song D. Aspect-based sentiment classification with aspect-specific graph convolutional networks[J]. arXiv preprint arXiv:1909.03477, 2019.
\bibitem{61} Zeng B, Yang H, Xu R, et al. Lcf: A local context focus mechanism for aspect-based sentiment classification[J]. Applied Sciences, 2019, 9(16): 3389.
\bibitem{62} Wang Y, Qian S, Hu J, et al. Fake news detection via knowledge-driven multimodal graph convolutional networks[C]//Proceedings of the 2020 International Conference on Multimedia Retrieval. 2020: 540-547.

\bibitem{63} Mesnil G, He X, Deng L, et al. Investigation of recurrent-neural-network architectures and learning methods for spoken language understanding[C]//Interspeech. 2013: 3771-3775.
\bibitem{64} Liu P, Joty S, Meng H. Fine-grained opinion mining with recurrent neural networks and word embeddings[C]//Proceedings of the 2015 conference on empirical methods in natural language processing. 2015: 1433-1443.
\bibitem{65} Mitchell M, Aguilar J, Wilson T, et al. Open domain targeted sentiment[C]//Proceedings of the 2013 Conference on Empirical Methods in Natural Language Processing. 2013: 1643-1654.
\bibitem{66} Cho K, Van Merrienboer B, Gulcehre C, et al. Learning phrase representations using RNN encoder-decoder for statistical machine translation[J]. arXiv preprint arXiv:1406.1078, 2014.
\bibitem{67} Sutskever I, Vinyals O, Le Q V. Sequence to sequence learning with neural networks[J]. Advances in neural information processing systems, 2014, 27.

\bibitem{68} Chen K, Wang J, Pang J, et al. MMDetection: Open mmlab detection toolbox and benchmark[J]. arXiv preprint arXiv:1906.07155, 2019.
\bibitem{69} Y. Wu, A. Kirillov, F. Massa, W.-Y. Lo, and R. Girshick, ''Detectron2,''https://github.com/facebookresearch/ detectron2, 2019.
\bibitem{70} Yang X, Feng S, Zhang Y, et al. Multimodal Sentiment Detection Based on Multi-channel Graph Neural Networks[C]//Proceedings of the 59th Annual Meeting of the Association for Computational Linguistics and the 11th International Joint Conference on Natural Language Processing (Volume 1: Long Papers). 2021: 328-339.
\bibitem{71} Amos B, Ludwiczuk B, Satyanarayanan M. Openface: A general-purpose face recognition library with mobile applications[J]. CMU School of Computer Science, 2016, 6(2): 20.
\bibitem{72} Wu, Yang \& Zhao, Yanyan \& Lu, Xin \& Qin, Bing \& Wu, Yin \& Sheng, Jian \& Li, Jinlong. (2021). Modeling Incongruity between Modalities for Multimodal Sarcasm Detection. IEEE MultiMedia. PP. 1-1. 10.1109/MMUL.2021.3069097. 
\bibitem{73} Tian Y L, Kanade T, Cohn J F. Facial expression analysis[M]//Handbook of face recognition. Springer, New York, NY, 2005: 247-275.

\bibitem{74} Tsai Y H H, Bai S, Liang P P, et al. Multimodal transformer for unaligned multimodal language sequences[C] //Proceedings of the conference. Association for Computational Linguistics. Meeting. NIH Public Access, 2019, 2019: 6558.
\bibitem{75} Degottex G, Kane J, Drugman T, et al. COVAREP-A collaborative voice analysis repository for speech technologies[C]//2014 ieee international conference on acoustics, speech and signal processing (icassp). IEEE, 2014: 960-964.
\bibitem{76} Wenmeng Yu, Hua Xu, Fanyang Meng, Yilin Zhu, Yixiao Ma, Jiele Wu, Jiyun Zou and Kaicheng Yang. CH-SIMS: A Chinese Multimodal Sentiment Analysis Dataset with Fine-grained Annotations of Modality. Proceedings of the 58th Annual Meeting of the Association for Computational Linguistics. 2020: 3718-3727.

\bibitem{77}Nan Xu, Wenji Mao, Guandan Chen. Multi-interactive memory network for aspect based multimodal sentiment analysis. Proceedings of the AAAI Conference on Artificial Intelligence. 2019: 371-378.

\bibitem{78} Yu J, Jiang J, Xia R. Entity-sensitive attention and fusion network for entity-level multimodal sentiment classification[J]. IEEE/ACM Transactions on Audio, Speech, and Language Processing, 2019, 28: 429-439.

\bibitem{79} Yu J, Jiang J. Adapting BERT for target-oriented multimodal sentiment classification[C]. IJCAI, 2019.

\bibitem{80} Ling Y, Yu J, Xia R. Vision-language pre-training for multimodal aspect-based sentiment analysis[J]. arXiv preprint arXiv:2204.07955, 2022.

\bibitem{81} Hu J, Liu Y, Zhao J, et al. MMGCN: Multimodal fusion via deep graph convolution network for emotion recognition in conversation[J]. arXiv preprint arXiv:2107.06779, 2021.
\bibitem{82} Wang Z, Ji H. Open Vocabulary Electroencephalography-To-Text Decoding and Zero-shot Sentiment Classification[J]. arXiv preprint arXiv:2112.02690, 2021.
\bibitem{83} Vinyals O, Toshev A, Bengio S, et al. Show and tell: A neural image caption generator[C]//Proceedings of the IEEE conference on computer vision and pattern recognition. 2015: 3156-3164.
\bibitem{84} Xu K, Ba J, Kiros R, et al. Show, attend and tell: Neural image caption generation with visual attention[C]//International conference on machine learning. PMLR, 2015: 2048-2057.
\bibitem{85} Chen, Long, et al. "SCA-CNN: Spatial and Channel-wise Attention in Convolutional Networks for Image Captioning." (2016):6298-6306.
\bibitem{86} P. Anderson, X. He, C. Buehler, D. Teney, M. Johnson, S. Gould, and L. Zhang. Bottom-up and top-down attention for image captioning and visual question answering. In CVPR, volume 3, page 6, 2018.
\bibitem{87} Zhou H, Huang M, Zhang T, et al. Emotional chatting machine: Emotional conversation generation with internal and external memory[C]//Proceedings of the AAAI Conference on Artificial Intelligence. 2018, 32(1).
\bibitem{88} Nezami O M, Dras M, Wan S, et al. Senti-attend: image captioning using sentiment and attention[J]. arXiv preprint arXiv:1811.09789, 2018.
\bibitem{89} Yu C, Lu H, Hu N, et al. Durian: Duration informed attention network for multimodal synthesis[J]. arXiv preprint arXiv:1909.01700, 2019.
\bibitem{90} Lewis M, Liu Y, Goyal N, et al. Bart: Denoising sequence-to-sequence pre-training for natural language generation, translation, and comprehension[J]. arXiv preprint arXiv:1910.13461, 2019.
\bibitem{91} Wu, Yang \& Zhao, Yanyan \& Lu, Xin \& Qin, Bing \& Wu, Yin \& Sheng, Jian \& Li, Jinlong. (2021). Modeling Incongruity between Modalities for Multimodal Sarcasm Detection. IEEE MultiMedia. PP. 1-1. 10.1109/MMUL.2021.3069097. 
\bibitem{92} J. Guo, J. Tang, W. Dai, Y. Ding, and W. Kong, ``Dynamically adjust word representations using unaligned multimodal information,'' in {\it Proc. the 30th ACM International Conference on Multimedia (MM)}, Lisbon, Portugal, 2022, pp. 3394-3402.
\bibitem{93} J. Tang, K. Li, X. Jin, A. Cichocki, Q. Zhao, and W. Kong, ``CTFN: hierarchical learning for multimodal sentiment analysis using coupled-translation fusion network,'' in {\it Proc. the 59th Annual Meeting of the Association for Computational Linguistics (ACL)}, Virtual, 2021, pp. 5301-5311.
\bibitem{94} A. Joshi, A. Bhat, A. Jain, A. V. Singh, and A. Modi, ``COGMEN: contextualized GNN based multimodal emotion recognition'', in {\it Proc. Conference of the North American Chapter of the Association for Computational Linguistics: Human Language Technologies (NAACL)}, Seattle, WA, 2022, pp. 4148-4164.
\bibitem{95} W. Yu, H. Xu, F. Meng, Y. Zhu, Y. Ma, J. Wu, J. Zou, and K. Yang, ``CH-SIMS: A Chinese multimodal sentiment analysis dataset with fine-grained annotation of modality,'' in {\it Proc. the 58th Annual Meeting of the Association for Computational Linguistics (ACL)}, Virtual, 2020, pp. 3718-3727.
\bibitem{96} Zheng Z, Zhang Z, Wang Z, Fu R, Liu M, Wang Z, Qin B. Decompose, Prioritize, and Eliminate: Dynamically Integrating Diverse Representations for Multimodal Named Entity Recognition. InProceedings of the 2024 Joint International Conference on Computational Linguistics, Language Resources and Evaluation (LREC-COLING 2024) 2024 May (pp. 4498-4508).
\bibitem{97} Yang X, Feng S, Zhang Y, et al. Multimodal sentiment detection based on multi-channel graph neural networks[C]// In {\it Proceedings of the 59th Annual Meeting of the Association for Computational Linguistics and the 11th International Joint Conference on Natural Language Processing (Volume 1: Long Papers)}. 2021: 328-339.
\bibitem{98} Yang H, Zhao Y, Qin B. Face-Sensitive Image-to-Emotional-Text Cross-modal Translation for Multimodal Aspect-based Sentiment Analysis[C]// In {\it Proceedings of the 2022 Conference on Empirical Methods in Natural Language Processing.} 2022: 3324-3335.
\bibitem{99} Amir Zadeh, Rowan Zellers, Eli Pincus, and LouisPhilippe Morency. 2016a. Mosi: Multimodal corpus of sentiment intensity and subjectivity analysis in online opinion videos. arXiv preprint arXiv:1606.06259.
\bibitem{100} AmirAli Bagher Zadeh, Paul Pu Liang, Soujanya Poria, Erik Cambria, and Louis-Philippe Morency. 2018b. Multimodal language analysis in the wild: Cmu-mosei dataset and interpretable dynamic fusion graph. In {\it Proceedings ofthe 56th Annual Meeting ofthe Association for Computational Linguistics (Volume 1: Long Papers)}, pages 2236-2246.
\bibitem{101} Wenmeng Yu, Hua Xu, Fanyang Meng, Yilin Zhu, Yixiao Ma, Jiele Wu, Jiyun Zou and Kaicheng Yang. CH-SIMS: A Chinese Multimodal Sentiment Analysis Dataset with Fine-grained Annotations of Modality. In {\it Proceedings of the 58th Annual Meeting of the Association for Computational Linguistics.} 2020: 3718-3727.
\bibitem{102} http://yelp.com/dataset.
\bibitem{103} T. Niu, S. A. Zhu, L. Pang and A. El Saddik, Sentiment Analysis on Multi-view Social Data, MultiMedia Modeling (MMM), pp: 15-27, Miami, 2016.
\bibitem{104} Yu J, Jiang J. Adapting BERT for target-oriented multimodal sentiment classification[C]. {\it IJCAI}, 2019.
\bibitem{105} Vaswani A, Shazeer N, Parmar N, et al. Attention is all you need[J]. {\it Advances in neural information processing systems}, 2017, 30.
\bibitem{106} Chen K, Wang J, Pang J, et al. MMDetection: Open mmlab detection toolbox and benchmark[J]. arXiv preprint arXiv:1906.07155, 2019.
\bibitem{107} Y. Wu, A. Kirillov, F. Massa, W.-Y. Lo, and R. Girshick, ``Detectron2,'' https://github.com/facebookresearch/ detectron2, 2019.
\bibitem{108} Zhao, Weixiang, Yanyan Zhao, Xin Lu, Shilong Wang, Yanpeng Tong, and Bing Qin. "Is ChatGPT Equipped with Emotional Dialogue Capabilities?." arXiv preprint arXiv:2304.09582 (2023).
\bibitem{109} Soleymani M, Garcia D, Jou B, Schuller B, Chang SF, Pantic M. A survey of multimodal sentiment analysis. Image and Vision Computing. 2017 Sep 1;65:3-14.
\bibitem{110} Girdhar R, El-Nouby A, Liu Z, Singh M, Alwala KV, Joulin A, Misra I. Imagebind: One embedding space to bind them all. InProceedings of the IEEE/CVF Conference on Computer Vision and Pattern Recognition 2023 (pp. 15180-15190).
\bibitem{111} Team G, Anil R, Borgeaud S, Wu Y, Alayrac JB, Yu J, Soricut R, Schalkwyk J, Dai AM, Hauth A, Millican K. Gemini: a family of highly capable multimodal models. arXiv preprint arXiv:2312.11805. 2023 Dec 19.
\bibitem{112} Li J, Li D, Savarese S, Hoi S. Blip-2: Bootstrapping language-image pre-training with frozen image encoders and large language models. InInternational conference on machine learning 2023 Jul 3 (pp. 19730-19742). PMLR.
\bibitem{113} Liu H, Li C, Wu Q, Lee YJ. Visual instruction tuning. Advances in neural information processing systems. 2024 Feb 13;36.
\bibitem{114} Bai J, Bai S, Yang S, Wang S, Tan S, Wang P, Lin J, Zhou C, Zhou J. Qwen-vl: A frontier large vision-language model with versatile abilities. arXiv preprint arXiv:2308.12966. 2023 Aug 24.
\bibitem{115} Chiang WL, Li Z, Lin Z, Sheng Y, Wu Z, Zhang H, Zheng L, Zhuang S, Zhuang Y, Gonzalez JE, Stoica I. Vicuna: An open-source chatbot impressing gpt-4 with 90\%* chatgpt quality. See https://vicuna. lmsys. org (accessed 14 April 2023). 2023 Mar;2(3):6.
\bibitem{116} Radford A, Kim JW, Hallacy C, Ramesh A, Goh G, Agarwal S, Sastry G, Askell A, Mishkin P, Clark J, Krueger G. Learning transferable visual models from natural language supervision. InInternational conference on machine learning 2021 Jul 1 (pp. 8748-8763). PMLR.
\bibitem{117} Lu P, Mishra S, Xia T, Qiu L, Chang KW, Zhu SC, Tafjord O, Clark P, Kalyan A. Learn to explain: Multimodal reasoning via thought chains for science question answering. Advances in Neural Information Processing Systems. 2022 Dec 6;35:2507-21.
\bibitem{118} Sharma P, Ding N, Goodman S, Soricut R. Conceptual captions: A cleaned, hypernymed, image alt-text dataset for automatic image captioning. InProceedings of the 56th Annual Meeting of the Association for Computational Linguistics (Volume 1: Long Papers) 2018 Jul (pp. 2556-2565).
\bibitem{119} Zhao H, Yang M, Bai X, Liu H. A Survey on Multimodal Aspect-Based Sentiment Analysis. IEEE Access. 2024 Jan 16.
\bibitem{120} Cai Y, Cai H, Wan X. Multi-modal sarcasm detection in twitter with hierarchical fusion model. InProceedings of the 57th annual meeting of the association for computational linguistics 2019 Jul (pp. 2506-2515).
\bibitem{121} Niu T, Zhu S, Pang L, El Saddik A. Sentiment analysis on multi-view social data. InMultiMedia Modeling: 22nd International Conference, MMM 2016, Miami, FL, USA, January 4-6, 2016, Proceedings, Part II 22 2016 (pp. 15-27). Springer International Publishing.
\bibitem{122} Poria S, Hazarika D, Majumder N, Naik G, Cambria E, Mihalcea R. Meld: A multimodal multi-party dataset for emotion recognition in conversations. arXiv preprint arXiv:1810.02508. 2018 Oct 5.
\bibitem{123} Ramamoorthy S, Gunti N, Mishra S, Suryavardan S, Reganti A, Patwa P, DaS A, Chakraborty T, Sheth A, Ekbal A, Ahuja C. Memotion 2: Dataset on sentiment and emotion analysis of memes. InProceedings of De-Factify: Workshop on Multimodal Fact Checking and Hate Speech Detection, CEUR 2022.
\bibitem{124} Castro S, Hazarika D, Pérez-Rosas V, Zimmermann R, Mihalcea R, Poria S. Towards multimodal sarcasm detection (an obviously perfect paper). arXiv preprint arXiv:1906.01815. 2019 Jun 5.
\bibitem{125} Zadeh A, Cao YS, Hessner S, Liang PP, Poria S, Morency LP. CMU-MOSEAS: A multimodal language dataset for Spanish, Portuguese, German and French. InProceedings of the Conference on Empirical Methods in Natural Language Processing. Conference on Empirical Methods in Natural Language Processing 2020 Nov (Vol. 2020, p. 1801). NIH Public Access.
\bibitem{126} Busso C, Bulut M, Lee CC, Kazemzadeh A, Mower E, Kim S, Chang JN, Lee S, Narayanan SS. IEMOCAP: Interactive emotional dyadic motion capture database. Language resources and evaluation. 2008 Dec;42:335-59.
\bibitem{127} Yang X, Feng S, Wang D, Zhang Y. Image-text multimodal emotion classification via multi-view attentional network. IEEE Transactions on Multimedia. 2020 Nov 2;23:4014-26.
\bibitem{128} Cai Y, Cai H, Wan X. Multi-modal sarcasm detection in twitter with hierarchical fusion model. InProceedings of the 57th annual meeting of the association for computational linguistics 2019 Jul (pp. 2506-2515).
\bibitem{129} M. Thelwall, K. Buckley, G. Paltoglou, D. Cai, and A. Kappas, "Sentiment strength detection in short informal text," Journal of the American Society for Information Science and Technology, vol. 61, pp. 2544-2558, 2010. 
\bibitem{130} D. Borth, R. Ji, T. Chen, T. Breuel, and S.-F. Chang, "Large-scale visual sentiment ontology and detectors using adjective noun pairs," in Proceedings of the 21st ACM international conference on Multimedia, 2013, pp. 223-232.
\bibitem{131} M. Wang, D. Cao, L. Li, S. Li, and R. Ji, "Microblog sentiment analysis based on cross-media bag-of-words model," in Proceedings of international conference on internet multimedia computing and service, 2014, p. 76. 
\bibitem{132} G. Cai and B. Xia, "Convolutional Neural Networks for Multimedia Sentiment Analysis," in National CCF Conference on Natural Language Processing and Chinese Computing, 2015, pp. 159-167.
\bibitem{133} Y. Yu, H. Lin, J. Meng, and Z. Zhao, "Visual and Textual Sentiment Analysis of a Microblog Using Deep Convolutional Neural Networks," Algorithms, vol. 9, p. 41, 2016.
\bibitem{134} Xu N. Analyzing multimodal public sentiment based on hierarchical semantic attentional network. In2017 IEEE international conference on intelligence and security informatics (ISI) 2017 Jul 22 (pp. 152-154). IEEE.
\bibitem{135} Xu N, Mao W. Multisentinet: A deep semantic network for multimodal sentiment analysis. InProceedings of the 2017 ACM on Conference on Information and Knowledge Management 2017 Nov 6 (pp. 2399-2402).
\bibitem{136} Xu N, Mao W, Chen G. A co-memory network for multimodal sentiment analysis. InThe 41st international ACM SIGIR conference on research \& development in information retrieval 2018 Jun 27 (pp. 929-932).
\bibitem{137} Yang X, Feng S, Zhang Y, Wang D. Multimodal sentiment detection based on multi-channel graph neural networks. InProceedings of the 59th Annual Meeting of the Association for Computational Linguistics and the 11th International Joint Conference on Natural Language Processing (Volume 1: Long Papers) 2021 Aug (pp. 328-339).
\bibitem{138} Li, Z., Xu, B., Zhu, C., \& Zhao, T. (2022). CLMLF: A contrastive learning and multi-layer fusion method for multimodal sentiment detection. arXiv preprint arXiv:2204.05515.
\bibitem{139} Tong Zhu, Leida Li, Jufeng Yang, Sicheng Zhao, Xiao Xiao, Multimodal emotion classification with multi-level semantic reasoning network, IEEE Trans. Multim. (2022) http://dx.doi.org/10.1109/TMM.2022.3214989, Early Access.
\bibitem{140} Jia A, He Y, Zhang Y, Uprety S, Song D, Lioma C. Beyond emotion: A multi-modal dataset for human desire understanding. InProceedings of the 2022 Conference of the North American Chapter of the Association for Computational Linguistics: Human Language Technologies 2022 Jul (pp. 1512-1522).
\bibitem{141} Wang W, Ding L, Shen L, Luo Y, Hu H, Tao D. WisdoM: Improving Multimodal Sentiment Analysis by Fusing Contextual World Knowledge. arXiv preprint arXiv:2401.06659. 2024 Jan 12.
\bibitem{142} Ma D, Li S, Wu F, Xie X, Wang H. Exploring sequence-to-sequence learning in aspect term extraction. InProceedings of the 57th annual meeting of the association for computational linguistics 2019 Jul (pp. 3538-3547).
\bibitem{143} Chen Z, Qian T. Enhancing aspect term extraction with soft prototypes. InProceedings of the 2020 Conference on Empirical Methods in Natural Language Processing (EMNLP) 2020 Nov (pp. 2107-2117).
\bibitem{144} Karamanolakis G, Hsu D, Gravano L. Leveraging just a few keywords for fine-grained aspect detection through weakly supervised co-training. arXiv preprint arXiv:1909.00415. 2019 Sep 1.
\bibitem{145} Z. Wu, C. Zheng, Y. Cai, J. Chen, H.-f. Leung, Q. Li, Multimodal representation with embedded visual guiding objects for named entity recognition in social media posts, in: Proceedings of the 28th ACM International Conference on Multimedia, 2020, pp. 1038–1046.
\bibitem{146} Q. Zhang, J. Fu, X. Liu, X. Huang, Adaptive co-attention network for named entity recognition in tweets, in: Proceedings of the AAAI Conference on Artificial Intelligence, Vol. 32, 2018.
\bibitem{147} L. Sun, J. Wang, K. Zhang, Y. Su, F. Weng, RpBERT: a text-image relation propagation-based BERT model for multimodal NER, in: Proceedings of the AAAI Conference on Artificial Intelligence, Vol. 35, 2021, pp. 13860–13868.
\bibitem{148} Yu J, Jiang J, Yang L, Xia R. Improving multimodal named entity recognition via entity span detection with unified multimodal transformer. Association for Computational Linguistics.
\bibitem{149} Moon S, Neves L, Carvalho V. Multimodal named entity recognition for short social media posts. arXiv preprint arXiv:1802.07862. 2018 Feb 22.
\bibitem{150} Zhang D, Wei S, Li S, Wu H, Zhu Q, Zhou G. Multi-modal graph fusion for named entity recognition with targeted visual guidance. InProceedings of the AAAI conference on artificial intelligence 2021 May 18 (Vol. 35, No. 16, pp. 14347-14355).
\bibitem{151} Peng T, Li Z, Wang P, Zhang L, Zhao H. A Novel Energy Based Model Mechanism for Multi-Modal Aspect-Based Sentiment Analysis. InProceedings of the AAAI Conference on Artificial Intelligence 2024 Mar 24 (Vol. 38, No. 17, pp. 18869-18878).
\bibitem{152} Peng T, Li Z, Zhang L, Du B, Zhao H. FSUIE: A Novel Fuzzy Span Mechanism for Universal Information Extraction. arXiv preprint arXiv:2306.14913. 2023 Jun 19.
\bibitem{153} Sundararaman MN, Ahmad Z, Ekbal A, Bhattacharyya P. Unsupervised aspect-level sentiment controllable style transfer. InProceedings of the 1st Conference of the Asia-Pacific Chapter of the Association for Computational Linguistics and the 10th International Joint Conference on Natural Language Processing 2020 Dec (pp. 303-312).
\bibitem{154} Ji Y, Liu H, He B, Xiao X, Wu H, Yu Y. Diversified multiple instance learning for document-level multi-aspect sentiment classification. InProceedings of the 2020 conference on empirical methods in natural language processing (EMNLP) 2020 Nov (pp. 7012-7023).
\bibitem{155} Liang B, Yin R, Gui L, Du J, Xu R. Jointly learning aspect-focused and inter-aspect relations with graph convolutional networks for aspect sentiment analysis. InProceedings of the 28th international conference on computational linguistics 2020 Dec (pp. 150-161).
\bibitem{156} Wei J, Bosma M, Zhao VY, Guu K, Yu AW, Lester B, Du N, Dai AM, Le QV. Finetuned language models are zero-shot learners. arXiv preprint arXiv:2109.01652. 2021 Sep 3.
\bibitem{157} Ouyang L, Wu J, Jiang X, Almeida D, Wainwright C, Mishkin P, Zhang C, Agarwal S, Slama K, Ray A, Schulman J. Training language models to follow instructions with human feedback. Advances in neural information processing systems. 2022 Dec 6;35:27730-44.
\bibitem{158} Wang Y, Kordi Y, Mishra S, Liu A, Smith NA, Khashabi D, Hajishirzi H. Self-instruct: Aligning language models with self-generated instructions. arXiv preprint arXiv:2212.10560. 2022 Dec 20.
\bibitem{159} Dai W, Li J, Li D, Tiong AM, Zhao J, Wang W, Li B, Fung PN, Hoi S. Instructblip: Towards general-purpose vision-language models with instruction tuning. Advances in Neural Information Processing Systems. 2024 Feb 13;36.
\bibitem{160} Feng J, Lin M, Shang L, Gao X. Autonomous Aspect-Image Instruction a2II: Q-Former Guided Multimodal Sentiment Classification. InProceedings of the 2024 Joint International Conference on Computational Linguistics, Language Resources and Evaluation (LREC-COLING 2024) 2024 May (pp. 1996-2005).
\bibitem{161} Ju X, Zhang D, Xiao R, Li J, Li S, Zhang M, Zhou G. Joint multi-modal aspect-sentiment analysis with auxiliary cross-modal relation detection. InProceedings of the 2021 conference on empirical methods in natural language processing 2021 Nov (pp. 4395-4405).
\bibitem{162} Yang L, Na JC, Yu J. Cross-modal multitask transformer for end-to-end multimodal aspect-based sentiment analysis. Information Processing \& Management. 2022 Sep 1;59(5):103038.
\bibitem{163} Xiao L, Wu X, Xu J, Li W, Jin C, He L. Atlantis: Aesthetic-oriented multiple granularities fusion network for joint multimodal aspect-based sentiment analysis. Information Fusion. 2024 Feb 15:102304.
\bibitem{164} Zhao F, Li C, Wu Z, Ouyang Y, Zhang J, Dai X. M2DF: Multi-grained Multi-curriculum Denoising Framework for Multimodal Aspect-based Sentiment Analysis. arXiv preprint arXiv:2310.14605. 2023 Oct 23.
\bibitem{165} Zhou R, Guo W, Liu X, Yu S, Zhang Y, Yuan X. AoM: Detecting aspect-oriented information for multimodal aspect-based sentiment analysis. arXiv preprint arXiv:2306.01004. 2023 May 31.
\bibitem{166} Liu Y, Zhou Y, Li Z, Zhang J, Shang Y, Zhang C, Hu S. RNG: Reducing Multi-level Noise and Multi-grained Semantic Gap for Joint Multimodal Aspect-Sentiment Analysis. arXiv preprint arXiv:2405.13059. 2024 May 20.
\bibitem{167} Li Y, Ding H, Lin Y, Feng X, Chang L. Multi-level textual-visual alignment and fusion network for multimodal aspect-based sentiment analysis. Artificial Intelligence Review. 2024 Apr;57(4):1-26.
\bibitem{168} Yu Y, Zhang D, Li S. Unified multi-modal pre-training for few-shot sentiment analysis with prompt-based learning. InProceedings of the 30th ACM International Conference on Multimedia 2022 Oct 10 (pp. 189-198).
\bibitem{169} Yu Y, Zhang D. Few-shot multi-modal sentiment analysis with prompt-based vision-aware language modeling. In2022 IEEE International Conference on Multimedia and Expo (ICME) 2022 Jul 18 (pp. 1-6). IEEE.
\bibitem{170} Yang X, Feng S, Wang D, Qi S, Wu W, Zhang Y, Hong P, Poria S. Few-shot joint multimodal aspect-sentiment analysis based on generative multimodal prompt. arXiv preprint arXiv:2305.10169. 2023 May 17.
\bibitem{171} Yang L, Wang Z, Li Z, Na JC, Yu J. An empirical study of Multimodal Entity-Based Sentiment Analysis with ChatGPT: Improving in-context learning via entity-aware contrastive learning. Information Processing \& Management. 2024 Jul 1;61(4):103724.
\bibitem{172} Morency LP, Mihalcea R, Doshi P. Towards multimodal sentiment analysis: Harvesting opinions from the web. InProceedings of the 13th international conference on multimodal interfaces 2011 Nov 14 (pp. 169-176).
\bibitem{173} Liu Y, Yuan Z, Mao H, Liang Z, Yang W, Qiu Y, Cheng T, Li X, Xu H, Gao K. Make acoustic and visual cues matter: CH-SIMS v2. 0 dataset and AV-Mixup consistent module. InProceedings of the 2022 International Conference on Multimodal Interaction 2022 Nov 7 (pp. 247-258).
\bibitem{174} Zhao J, Zhang T, Hu J, Liu Y, Jin Q, Wang X, Li H. M3ED: Multi-modal multi-scene multi-label emotional dialogue database. arXiv preprint arXiv:2205.10237. 2022 May 9.
\bibitem{175} Lian Z, Sun H, Sun L, Chen K, Xu M, Wang K, Xu K, He Y, Li Y, Zhao J, Liu Y. Mer 2023: Multi-label learning, modality robustness, and semi-supervised learning. InProceedings of the 31st ACM International Conference on Multimedia 2023 Oct 26 (pp. 9610-9614).
\bibitem{176} Lian Z, Sun H, Sun L, Wen Z, Zhang S, Chen S, Gu H, Zhao J, Ma Z, Chen X, Yi J. MER 2024: Semi-Supervised Learning, Noise Robustness, and Open-Vocabulary Multimodal Emotion Recognition. arXiv preprint arXiv:2404.17113. 2024 Apr 26.
\bibitem{177} Lian Z, Sun L, Xu M, Sun H, Xu K, Wen Z, Chen S, Liu B, Tao J. Explainable multimodal emotion reasoning. arXiv preprint arXiv:2306.15401. 2023 Jun 27.
\bibitem{178} Hasan MK, Rahman W, Zadeh A, Zhong J, Tanveer MI, Morency LP. UR-FUNNY: A multimodal language dataset for understanding humor. arXiv preprint arXiv:1904.06618. 2019 Apr 14.
\bibitem{179} Wei X, Zhang T, Li Y, Zhang Y, Wu F. Multi-modality cross attention network for image and sentence matching. InProceedings of the IEEE/CVF conference on computer vision and pattern recognition 2020 (pp. 10941-10950).
\bibitem{180} Anderson P, He X, Buehler C, Teney D, Johnson M, Gould S, Zhang L. Bottom-up and top-down attention for image captioning and visual question answering. InProceedings of the IEEE conference on computer vision and pattern recognition 2018 (pp. 6077-6086).
\bibitem{181} Huang J, Pu Y, Zhou D, Shi H, Zhao Z, Xu D, Cao J. Multimodal Sentiment Analysis Based on 3D Stereoscopic Attention. InICASSP 2024-2024 IEEE International Conference on Acoustics, Speech and Signal Processing (ICASSP) 2024 Apr 14 (pp. 11151-11155). IEEE.
\bibitem{182} Mai S, Zeng Y, Zheng S, Hu H. Hybrid contrastive learning of tri-modal representation for multimodal sentiment analysis. IEEE Transactions on Affective Computing. 2022 May 3.
\bibitem{183} Lin R, Hu H. Multimodal contrastive learning via uni-Modal coding and cross-Modal prediction for multimodal sentiment analysis. arXiv preprint arXiv:2210.14556. 2022 Oct 26.
\bibitem{184} Sun Z, Sarma P, Sethares W, Liang Y. Learning relationships between text, audio, and video via deep canonical correlation for multimodal language analysis. InProceedings of the AAAI Conference on Artificial Intelligence 2020 Apr 3 (Vol. 34, No. 05, pp. 8992-8999).
\bibitem{185} Hazarika D, Zimmermann R, Poria S. Misa: Modality-invariant and-specific representations for multimodal sentiment analysis. InProceedings of the 28th ACM international conference on multimedia 2020 Oct 12 (pp. 1122-1131).
\bibitem{186} Mai S, Hu H, Xing S. Modality to modality translation: An adversarial representation learning and graph fusion network for multimodal fusion. InProceedings of the AAAI Conference on Artificial Intelligence 2020 Apr 3 (Vol. 34, No. 01, pp. 164-172).
\bibitem{187} Liu Z, Shen Y, Lakshminarasimhan VB, Liang PP, Zadeh A, Morency LP. Efficient low-rank multimodal fusion with modality-specific factors. arXiv preprint arXiv:1806.00064. 2018 May 31.
\bibitem{188} Mai S, Hu H, Xing S. Divide, conquer and combine: Hierarchical feature fusion network with local and global perspectives for multimodal affective computing. InProceedings of the 57th annual meeting of the association for computational linguistics 2019 Jul (pp. 481-492).
\bibitem{189} Chen M, Wang S, Liang PP, Baltrušaitis T, Zadeh A, Morency LP. Multimodal sentiment analysis with word-level fusion and reinforcement learning. InProceedings of the 19th ACM international conference on multimodal interaction 2017 Nov 3 (pp. 163-171).
\bibitem{190} Zadeh A, Liang PP, Mazumder N, Poria S, Cambria E, Morency LP. Memory fusion network for multi-view sequential learning. InProceedings of the AAAI conference on artificial intelligence 2018 Apr 27 (Vol. 32, No. 1).
\bibitem{191} Rahman W, Hasan MK, Lee S, Zadeh A, Mao C, Morency LP, Hoque E. Integrating multimodal information in large pretrained transformers. InProceedings of the conference. Association for Computational Linguistics. Meeting 2020 Jul (Vol. 2020, p. 2359). NIH Public Access.
\bibitem{192} Dai W, Cahyawijaya S, Liu Z, Fung P. Multimodal end-to-end sparse model for emotion recognition. arXiv preprint arXiv:2103.09666. 2021 Mar 17.
\bibitem{193} Morency LP, Mihalcea R, Doshi P. Towards multimodal sentiment analysis: Harvesting opinions from the web. InProceedings of the 13th international conference on multimodal interfaces 2011 Nov 14 (pp. 169-176).
\bibitem{194} Schifanella R, De Juan P, Tetreault J, Cao L. Detecting sarcasm in multimodal social platforms. InProceedings of the 24th ACM international conference on Multimedia 2016 Oct 1 (pp. 1136-1145).
\bibitem{195} H. Pan, Z. Lin, P. Fu, Y. Qi, W. Wang, Modeling intra and inter-modality incongruity for multi-modal sarcasm detection, in: Findings of the Association for Computational Linguistics, EMNLP 2020, 2020, pp. 1383–1392.
\bibitem{196} X. Wang, X. Sun, T. Yang, H. Wang, Building a bridge: A method for image-text sarcasm detection without pretraining on image-text data, in: Proceedings of the First International Workshop on Natural Language Processing beyond Text, 2020, pp. 19–29.
\bibitem{197} D. Tomás, R. Ortega-Bueno, G. Zhang, P. Rosso, R. Schifanella, Transformer-based models for multimodal irony detection, J. Ambient Intell. Humaniz. Comput. (2022) 1–12.
\bibitem{198} B. Liang, C. Lou, X. Li, L. Gui, M. Yang, R. Xu, Multi-modal sarcasm detection with interactive in-modal and cross-modal graphs, in: Proceedings of the 29th ACM International Conference on Multimedia, 2021, pp. 4707–4715.
\bibitem{199} B. Liang, C. Lou, X. Li, M. Yang, L. Gui, Y. He, W. Pei, R. Xu, Multi-modal sarcasm detection via cross-modal graph convolutional network, in: Proceedings of the 60th Annual Meeting of the Association for Computational Linguistics (Volume 1: Long Papers), 2022, pp. 1767–1777.
\bibitem{200} Yue T, Mao R, Wang H, Hu Z, Cambria E. KnowleNet: Knowledge fusion network for multimodal sarcasm detection. Information Fusion. 2023 Dec 1;100:101921.
\bibitem{201} Speer R, Chin J, Havasi C. Conceptnet 5.5: An open multilingual graph of general knowledge. InProceedings of the AAAI conference on artificial intelligence 2017 Feb 12 (Vol. 31, No. 1).
\bibitem{202} H. Liu, B. Yang, Z. Yu, A multi-view interactive approach for multimodal sarcasm detection in social internet of things with knowledge enhancement, Appl. Sci. 14 (5) (2024) 2146.
\bibitem{203} H. Fu, H. Liu, H. Wang, L. Xu, J. Lin, D. Jiang, Multi-modal sarcasm detection with sentiment word embedding, Electronics 13 (5) (2024) 855. 
\bibitem{204} Yi G, Fan C, Zhu K, Lv Z, Liang S, Wen Z, Pei G, Li T, Tao J. Vlp2msa: expanding vision-language pre-training to multimodal sentiment analysis. Knowledge-Based Systems. 2024 Jan 11;283:111136.
\bibitem{205} Qin L, Huang S, Chen Q, Cai C, Zhang Y, Liang B, Che W, Xu R. MMSD2. 0: Towards a Reliable Multi-modal Sarcasm Detection System. arXiv preprint arXiv:2307.07135. 2023 Jul 14.
\bibitem{206} Leveraging Generative Large Language Models with Visual Instruction and Demonstration Retrieval for Multimodal Sarcasm Detection. openreview Dec.2023. https://openreview.net/forum?id=\_98UHlfKejb
\bibitem{207} Lin H, Chen Z, Luo Z, Cheng M, Ma J, Chen G. CofiPara: A Coarse-to-fine Paradigm for Multimodal Sarcasm Target Identification with Large Multimodal Models. arXiv preprint arXiv:2405.00390. 2024 May 1.
\bibitem{208} Qin L, Chen Q, Feng X, Wu Y, Zhang Y, Li Y, Li M, Che W, Yu PS. Large Language Models Meet NLP: A Survey. arXiv preprint arXiv:2405.12819. 2024 May 21.
\bibitem{209} Houlsby N, Giurgiu A, Jastrzebski S, Morrone B, De Laroussilhe Q, Gesmundo A, Attariyan M, Gelly S. Parameter-efficient transfer learning for NLP. InInternational conference on machine learning 2019 May 24 (pp. 2790-2799). PMLR.
\bibitem{210} Hu EJ, Shen Y, Wallis P, Allen-Zhu Z, Li Y, Wang S, Wang L, Chen W. Lora: Low-rank adaptation of large language models. arXiv preprint arXiv:2106.09685. 2021 Jun 17.
\bibitem{211} Li XL, Liang P. Prefix-tuning: Optimizing continuous prompts for generation. arXiv preprint arXiv:2101.00190. 2021 Jan 1.
\bibitem{212} Dettmers T, Pagnoni A, Holtzman A, Zettlemoyer L. Qlora: Efficient finetuning of quantized llms. Advances in Neural Information Processing Systems. 2024 Feb 13;36.
\bibitem{213} Zhang Z, Peng L, Pang T, Han J, Zhao H, Schuller BW. Refashioning emotion recognition modelling: The advent of generalised large models. IEEE Transactions on Computational Social Systems. 2024 May 30.
\bibitem{214} Peng L, Zhang Z, Pang T, Han J, Zhao H, Chen H, Schuller BW. Customising General Large Language Models for Specialised Emotion Recognition Tasks. InICASSP 2024-2024 IEEE International Conference on Acoustics, Speech and Signal Processing (ICASSP) 2024 Apr 14 (pp. 11326-11330). IEEE.
\bibitem{215} Geetha AV, Mala T, Priyanka D, Uma E. Multimodal Emotion Recognition with deep learning: advancements, challenges, and future directions. Information Fusion. 2024 May 1;105:102218.
\bibitem{216} LeCun Y, Chopra S, Hadsell R, Ranzato M, Huang F. A tutorial on energy-based learning. Predicting structured data. 2006 Aug 19;1(0).
\bibitem{217} Yang X, Wu W, Feng S, Wang M, Wang D, Li Y, Sun Q, Zhang Y, Fu X, Poria S. MM-InstructEval: Zero-Shot Evaluation of (Multimodal) Large Language Models on Multimodal Reasoning Tasks. arXiv preprint arXiv:2405.07229. 2024 May 12.





\end{thebibliography}
\end{document}